\ifdefined\XeTeXrevision\else\pdfoutput=1\fi
\documentclass[preprint,5p,times,twocolumn,authoryear]{elsarticle}

\usepackage{amsmath,amssymb,amsthm}
\usepackage{graphicx}
\usepackage{booktabs}
\usepackage{array}
\usepackage{multirow}
\usepackage{placeins}
\usepackage{xcolor}
\usepackage{algorithm}
\usepackage{algpseudocode}
\usepackage[hidelinks]{hyperref}
\newcolumntype{L}[1]{>{\raggedright\arraybackslash}p{#1}}

\newtheorem{theorem}{Theorem}
\newtheorem{proposition}{Proposition}
\newtheorem{corollary}{Corollary}
\newtheorem{lemma}{Lemma}
\theoremstyle{remark}
\newtheorem{remark}{Remark}
\theoremstyle{definition}
\newtheorem{assumption}{Assumption}

\newcommand{\E}{\mathbb{E}}
\renewcommand{\P}{\mathbb{P}}
\newcommand{\1}{\mathbf{1}}
\newcommand{\FR}{\mathrm{FR}}
\newcommand{\Sens}{\mathrm{Sens}}
\newcommand{\dis}{\mathrm{dis}}
\newcommand{\all}{\mathrm{all}}
\newcommand{\clr}{\mathrm{clear}}
\newcommand{\pre}{\mathrm{pre}}

\makeatletter
\def\ps@pprintTitle{\let\@oddhead\@empty\let\@evenhead\@empty\let\@oddfoot\@empty\let\@evenfoot\@empty}
\makeatother

\begin{document}

\begin{frontmatter}

\title{When accuracy hides risk: risk-controlled confirmation of the critical view of safety under expert disagreement}

\author[vu,xs]{Bowen Xiang\corref{cor1}}
\ead{bowen.xiang@vanderbilt.edu}
\author[xs]{Xiao Yu}
\author[xs]{Mingze Xu}
\cortext[cor1]{Corresponding author.}
\affiliation[vu]{organization={Department of Biomedical Engineering, Vanderbilt University},
            addressline={2301 Vanderbilt Place}, city={Nashville}, state={TN}, postcode={37235}, country={USA}}
\affiliation[xs]{organization={Xiangshu MedTech}, city={Hangzhou}, country={China}}

\begin{abstract}
Automated assessment of the critical view of safety (CVS) in laparoscopic cholecystectomy has advanced rapidly, but a deployed check must decide, before the first clip, whether to confirm CVS. The standard rule calibrates a threshold so that false confirmations are rare on average. We show that this rule concentrates its errors on disputed videos, which one of three expert raters judged sufficient, and that the failure is structural. When reference labels are votes of a panel sampled from a larger expert pool, the rule that maximizes sensitivity at a fixed average false-confirmation rate also maximizes false confirmation of disputed videos, and for recognizers that recover expert judgement with less noise this rate rises with accuracy. We propose certified confirmation, a risk-controlled monitoring rule whose threshold is selected by multiple testing so that, with probability at least $1-\delta$, false confirmation is at most $\alpha$ overall, among disputed videos, within vendor and country groups, and before CVS is visible, using disputed labels only to calibrate. In a pre-specified, hash-locked evaluation on the 300 test videos (18 countries) of the SAGES CVS Challenge, standard calibration falsely confirmed 8 of 39 disputed videos (0.21; one-sided $p = 0.037$ against $\alpha = 0.10$), whereas certified confirmation confirmed 2 (one-sided 95\,\% upper bound 0.15). On development data, protection held for vendor and country groups left out of calibration. The cost was confirming 10\,\% rather than 35\,\% of sufficient videos on time. Average accuracy hides this risk; disputed cases must be protected explicitly.
\end{abstract}

\begin{keyword}
critical view of safety \sep surgical video analysis \sep rater disagreement \sep reference standard \sep risk control \sep conformal calibration \sep laparoscopic cholecystectomy
\end{keyword}

\end{frontmatter}

\section{Introduction}\label{sec:intro}

Bile duct injury remains the most feared complication of laparoscopic cholecystectomy. It is rare but causes lasting morbidity \citep{rystedt2016bile}, and most injuries follow a misidentification of anatomy rather than a technical slip \citep{way2003causes}. The critical view of safety (CVS) was introduced to prevent such misidentification \citep{strasberg1995analysis,strasberg2010rationale}. It has three criteria: a hepatocystic triangle cleared of fat and fibrous tissue (C1), the lower third of the gallbladder separated from the cystic plate (C2), and two, and only two, structures entering the gallbladder (C3). Guidelines ask surgeons to establish CVS before clipping \citep{brunt2020safe}, yet video reviews show that it is often not reached when surgeons believe it is \citep{nijssen2015complications}. An automated, independent check at the moment of clipping is therefore an obvious application of surgical data science \citep{maierhein2017surgical}.

\begin{figure*}[!t]
\centering
\includegraphics[width=\textwidth]{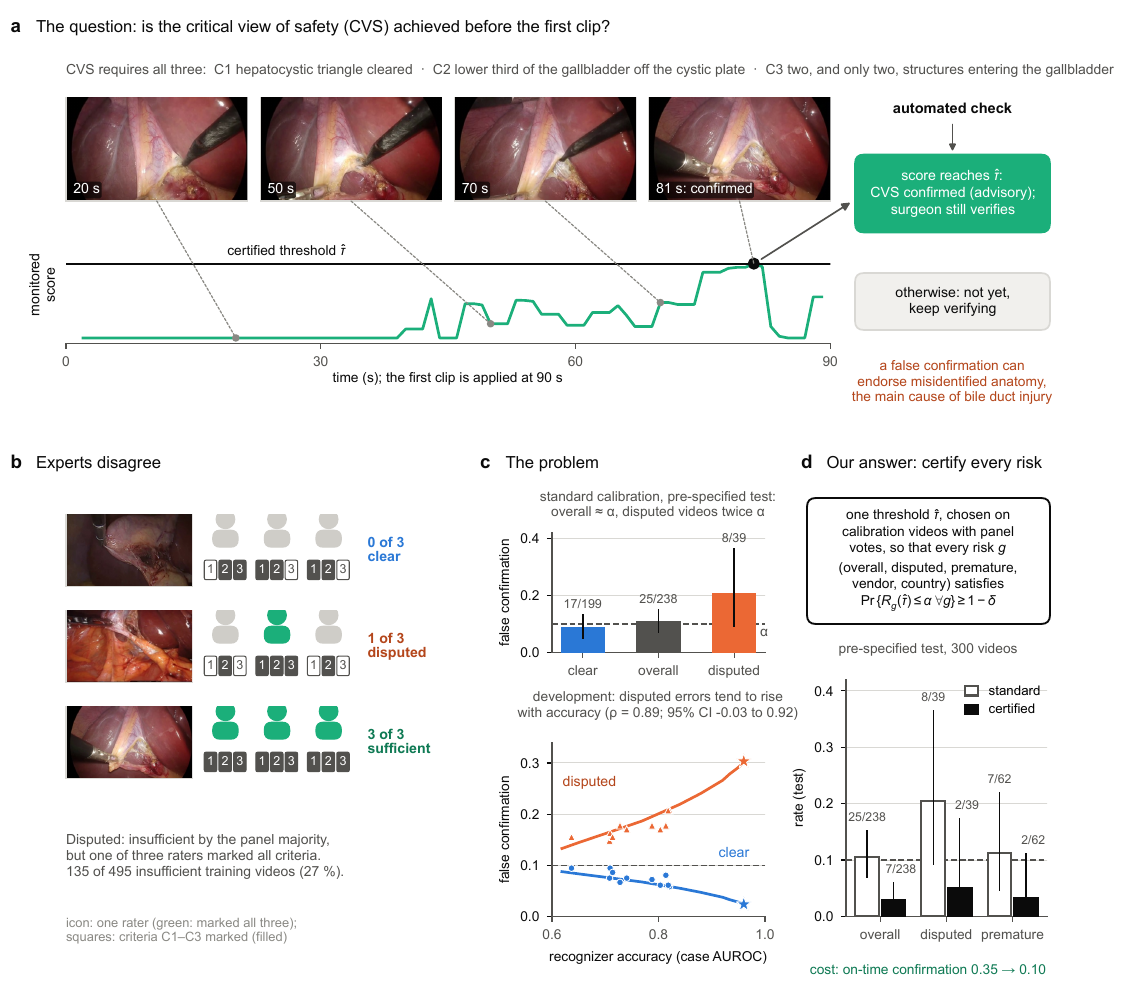}
\caption{The question, why it matters, and our answer.
(a) The question. Frames of a sufficient development video at 20, 50, 70 and 81\,s, tied to its monitored score (frame scorer; \ref{app:examples}). Before the first clip, an automated check either confirms CVS, here at 81\,s when the score reaches the certified threshold, or asks the surgeon to keep verifying. A confirmation is advisory: the surgeon verifies CVS independently before clipping, and a false confirmation is the error that can mislead.
(b) Experts disagree. Three development videos with each rater's video-level marks on C1--C3 (squares, filled if marked); a figure is green if that rater marked all three. Disputed videos are insufficient by majority, but one rater judged them sufficient.
(c) The problem. Top: under standard calibration, which keeps the \emph{average} false-confirmation rate at $\alpha = 0.1$ (dashed), false confirmations on the pre-specified test concentrate on disputed videos (frame scorer; exact 95\,\% CIs). Bottom: on development data, disputed false confirmation tends to rise with recognizer accuracy while that of clear videos falls (Spearman $\rho = 0.89$; bootstrap 95\,\% CI $-0.03$ to $0.92$). Lines: prediction of the panel model fitted to training votes (Section~\ref{sec:res-paradox}; model-based); points: 11 cross-fitted scorers; stars: the model's ideal recognizer.
(d) Our answer. Certified confirmation selects one threshold so that every certified risk $R_g$ is at most $\alpha$ with probability at least $1-\delta$ (Section~\ref{sec:method}); a confirmation is an advisory second check, not a replacement for the surgeon's own, and on test about half of all confirmations were of insufficient videos (Table~\ref{tab:readout}). Bars: false confirmation overall and among disputed videos, and premature confirmation, on the pre-specified test under standard calibration (open) and the certified rule (filled), with exact 95\,\% CIs; the cost is on-time confirmation of sufficient videos.
Frames: SAGES CVS Challenge 2024 training videos, CC BY-NC 4.0 \citep{alapatt2026sages}.}\label{fig:overview}
\end{figure*}

Recognition of the criteria in video has progressed quickly, through frame classifiers \citep{mascagni2022artificial,kawamura2023development}, anatomy-aware graph models \citep{murali2023latent} and public datasets \citep{murali2023endoscapes,rios2023cholec80cvs}. It culminated in the SAGES CVS Challenge \citep{alapatt2026sages}: 1,000 videos from 54 institutions in 24 countries, each ending at the first clip and rated by three experts drawn from a pool of twenty. The challenge ranks methods by frame-level criterion recognition, using mean average precision (mAP), calibration and robustness across centres, and leaves open how frame predictions should become decisions about a video or procedure \citep{alapatt2026sages}.

A deployed check has to take that decision. At the moment of clipping it either \emph{confirms} that CVS was achieved or asks the surgeon to verify (Fig.~\ref{fig:overview}a). A false confirmation is the error that could contribute to an injury. The standard way to control it is to calibrate the confirmation threshold on held-out cases, so that false confirmations stay below a level $\alpha$ on average. Conformal calibration does this with a finite-sample guarantee \citep{vovk2005algorithmic,angelopoulos2023gentle}.

An average guarantee can hide the cases that matter most \citep{oakdenrayner2020hidden}. CVS is judged by experts, and experts disagree (Fig.~\ref{fig:overview}b). In the SAGES training set, 135 of the 495 insufficient videos (27\,\%) were judged \emph{sufficient} by one of their three raters. We call these videos \emph{disputed}. A false confirmation of a disputed video is the failure an independent check is meant to prevent: a confirmation that the majority of an independent panel would reject. We find that standard calibration confirms disputed videos at about twice the nominal rate, and that, across our development scorers, more accurate recognizers tended to do worse (Fig.~\ref{fig:overview}c), an association that is uncertain but consistent with the theory below.

The question this paper addresses is therefore methodological: how should a confirmation rule be calibrated and validated when its reference labels are votes of small expert panels, so that the error rate it declares also holds where the panel splits? CVS is the motivating case, and errors are measured against the reference used to label CVS datasets, the per-criterion majorities of an independent expert panel; Section~\ref{sec:target} explains why this target must hold for disputed videos and not only on average. We do not claim that the resulting check is clinically useful as it stands: with current recognizers it confirms few sufficient videos, and its effect on surgeons and outcomes would need a prospective study (Section~\ref{sec:discussion}). The lesson, that accuracy and average error rates hide where errors fall when labels come from panels, applies wherever medical images are labelled this way.

This paper explains why this accuracy--safety paradox occurs, in which average accuracy hides where the risk falls, and provides a decision rule that removes it (Fig.~\ref{fig:overview}d). Our contributions are:
\begin{enumerate}
\item \textbf{A theory of the paradox} (Section~\ref{sec:theory}). When reference labels are votes of a panel drawn from a larger expert pool, the vote count has a monotone likelihood ratio in the video's latent expert share. We prove that, among all video-based rules with the same average false-confirmation rate, the rule that maximizes sensitivity also maximizes false confirmation of disputed videos (Theorem~\ref{thm:paradox}), and that this rate increases as recognizers recover expert judgement with less noise (Theorem~\ref{thm:monotone}). We bound what protection must cost (Proposition~\ref{prop:price}) and show that, in the same model, a recognizer held to the disputed limit gains sensitivity rather than disputed errors as its noise falls (Corollary~\ref{cor:gain}). We also show which part of the cost the data can identify, and give the minimum number of calibration cases any protected group needs (Lemma~\ref{lem:nmin}).
\item \textbf{Certified confirmation} (Section~\ref{sec:method}), a monitoring rule whose single threshold is selected by Learn-then-Test multiple testing \citep{angelopoulos2025learn}. With probability at least $1-\delta$ over the calibration set, it bounds false confirmation overall, among disputed videos, within vendor and country groups, and before CVS becomes visible. Its calibration machinery is standard group-wise high-probability risk control (Section~\ref{sec:relation}); what is new is what it certifies and why: a group that cannot be observed at the bedside and is needed only in the calibration set, a timing risk on a monitored stream, and the theory that makes these risks necessary.
\item \textbf{Pre-specified multi-centre evidence} (Sections~\ref{sec:design}--\ref{sec:results}). Development used cross-fitting on the 700 SAGES training videos. The protocol, code, models and thresholds were hash-locked before any test label was downloaded, and the 300 test videos were evaluated once. We report the paradox, the protection under left-out vendors and countries and on the test set, the clinical read-out, and the method variants that failed.
\end{enumerate}

\section{Related work}\label{sec:related}

\paragraph{Automated CVS assessment} Work on automated CVS assessment includes frame classification of the criteria \citep{mascagni2022artificial}, an annotation protocol for hepatocystic anatomy and CVS \citep{mascagni2021protocol}, anatomy-aware latent graphs \citep{murali2023latent} and real-time systems \citep{kawamura2023development}. Public datasets provide frame-level criteria from three experts: Endoscapes2023 \citep{murali2023endoscapes}, Cholec80-CVS \citep{rios2023cholec80cvs} on Cholec80 \citep{twinanda2017endonet}, and the multi-centre SAGES challenge \citep{alapatt2026sages}. These studies evaluate recognition, with average precision (AP), mAP and calibration error. To our knowledge, none provides a confirmation rule with a guarantee on false confirmation, or examines false confirmation where raters disagree.

\paragraph{Rater disagreement and imperfect reference standards} Learning from multiple annotators \citep{dawid1979maximum,raykar2010learning}, label noise in medical imaging \citep{karimi2020deep} and the use of inter-rater agreement to improve uncertainty estimates \citep{jensen2019improving} address disagreement during training. Grader variability changes measured diagnostic performance \citep{krause2018grader}, and imperfect reference standards are a long-standing problem in diagnostic accuracy studies \citep{reitsma2009review}. \citet{stutz2023evaluating} evaluate models against a distribution over plausible labels, and conformal prediction has been adapted to such ambiguous ground truth \citep{stutz2023conformal}. Hidden stratification, where aggregate metrics mask failures in clinically important subsets \citep{oakdenrayner2020hidden,sohoni2020nosubclass}, and the finding that selective classification can magnify disparities between groups \citep{jones2021selective} show that averages can mislead; validation guidelines warn of the same for rankings and metrics \citep{maierhein2018rankings,maierhein2024metrics,varoquaux2022machine}. We keep the panel majority as the reference, as clinical audit would, and contribute a mechanism: for decisions calibrated to an average risk, panel sampling makes the disputed stratum the one in which errors concentrate, and better recognition makes this worse.

\paragraph{Distribution-free risk control} Conformal prediction gives finite-sample guarantees in expectation over the calibration set \citep{vovk2005algorithmic,angelopoulos2023gentle}. High-probability versions, also called training-conditional or probably approximately correct (PAC), bound the realized risk with probability $1-\delta$: training-conditional conformal prediction \citep{vovk2012conditional,park2020pac,bian2023training}, Neyman--Pearson umbrella classification \citep{tong2018neyman} and selection with guaranteed risk \citep{geifman2017selective}. Risk-controlling prediction sets, conformal risk control and Learn-then-Test extend control to general and multiple risks \citep{bates2021distribution,angelopoulos2024conformal,angelopoulos2025learn}; Learn-then-Test has been used for clinical triage with abstention \citep{angelopoulos2024triage}. Guarantees within observed groups come from equalized and conditional coverage \citep{romano2020malice,gibbs2025conformal}, multivalid coverage \citep{jung2023batch}, fair risk control \citep{zhang2024fair} and fair conformal prediction in medical imaging \citep{lu2022fair}. These methods either control a risk in expectation or need group membership when the rule is applied. Our rule is a direct instance of the high-probability tools (Section~\ref{sec:relation}); what is new is the theory that says which group must be protected and why recognition cannot remove the need, the certification of a group observed only during calibration together with a timing risk on a monitored stream, and a pre-specified evaluation of the resulting system.

\paragraph{Deferral and decision support} Uncertainty-based referral \citep{leibig2017leveraging,kompa2021second} and learned deferral to clinicians \citep{dvijotham2023enhancing} decide when a model should act. Calibration of deep networks \citep{guo2017calibration} makes scores interpretable as probabilities. Automation bias \citep{goddard2012automation} is the main human-factors risk of a confirming system, which motivates conservative confirmation.

\section{Problem formulation}\label{sec:problem}

\begin{figure*}[p]
\centering
\includegraphics[width=\textwidth,height=0.77\textheight,keepaspectratio]{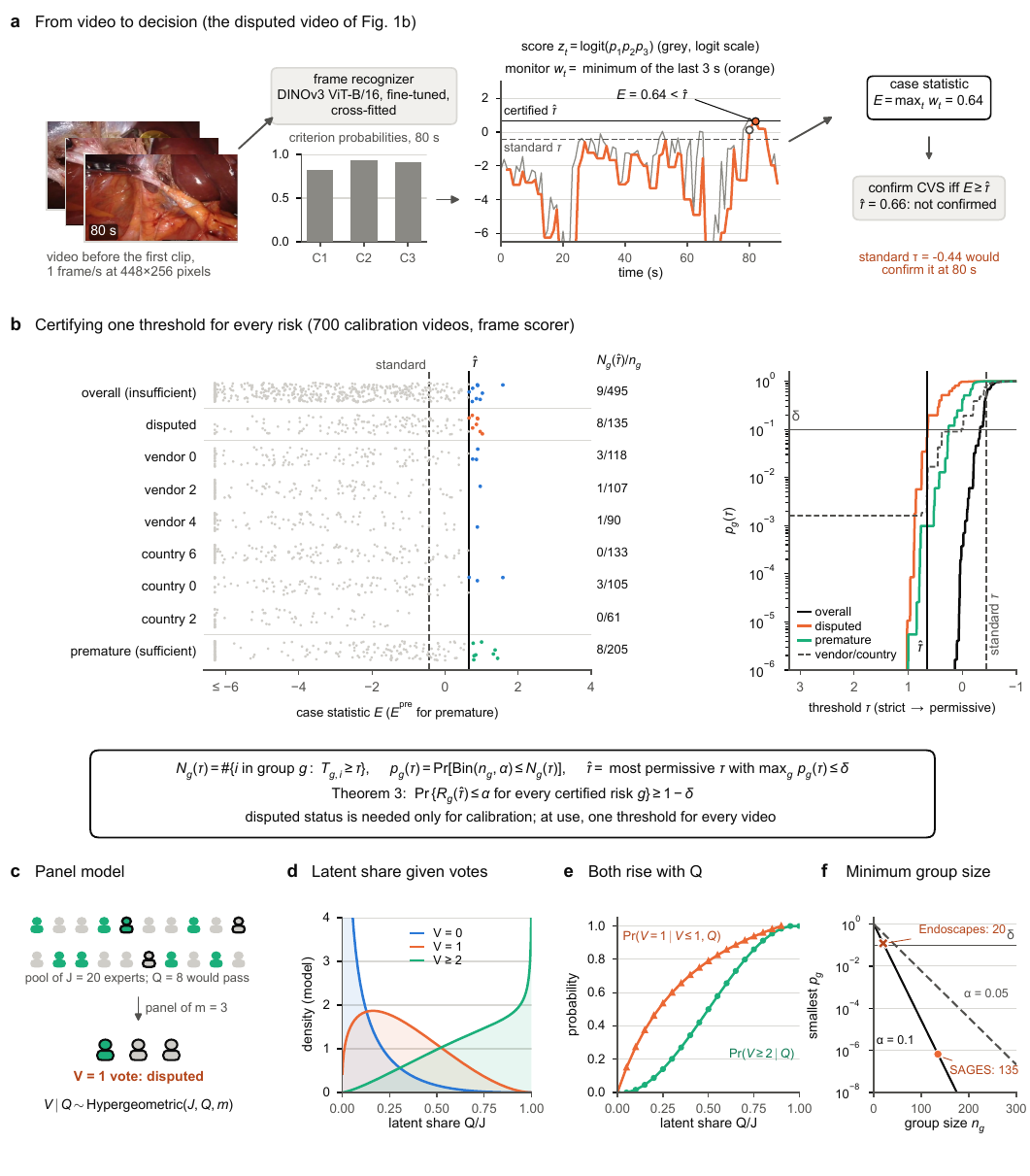}
\caption{Certified confirmation in detail.
(a) From video to decision, for the disputed video of Fig.~\ref{fig:overview}b. A cross-fitted frame recognizer gives criterion probabilities every second; the per-second score $z_t = \mathrm{logit}(p_1p_2p_3)$ (grey) is monitored as its 3-s minimum $w_t$ (orange), and the case statistic $E = \max_t w_t$ decides confirmation. The standard threshold would confirm this video at 80\,s; its peak $E = 0.64$ stays below the certified threshold $\hat\tau = 0.66$.
(b) Certifying one threshold on the 700 development videos (frame scorer). Left: the case statistic of every calibration video in each certified group (all insufficient videos, the disputed ones, the insufficient videos of each vendor and country group, and $E^{\mathrm{pre}}$ for sufficient videos), coloured if at or above $\hat\tau$, with the count $N_g(\hat\tau)$ of $n_g$. Right: the binomial $p$-value of each risk as the threshold is relaxed from strict to permissive (the axis is reversed); ``vendor/country'' is the largest of the six group $p$-values. Solid: $\hat\tau$; dashed: the standard threshold. $\hat\tau$ is the most permissive threshold at which every $p$-value is still at most $\delta$; here the disputed risk binds, and the next candidate, the peak of the video in (a), would fail.
(c) Panel model (Assumption~\ref{ass:panel}). A video has a latent number $Q$ of the $J = 20$ experts in the pool who would pass it (green). Its label comes from a panel of $m = 3$ of them, so the vote count $V$ is hypergeometric given $Q$; with one vote of three, this video is disputed.
(d) Latent expert share given the vote count, under the beta-binomial model fitted to SAGES training votes (model-based, continuous approximation; Section~\ref{sec:res-paradox}): videos with one vote lie between those with none and those with two or three.
(e) The probabilities of ``sufficient'' and of ``disputed among insufficient'' both increase with $Q$. This shared monotonicity is why the most sensitive rule at a fixed average level also confirms the most disputed videos (Theorem~\ref{thm:paradox}), and why recognition noise lowers both (Theorem~\ref{thm:monotone}).
(f) Lemma~\ref{lem:nmin}: with $\delta = 0.1$, a group with $n_g$ calibration videos can be certified only if $(1-\alpha)^{n_g} \le \delta$, that is, with at least 22 videos at $\alpha = 0.1$ (solid) and 45 at $\alpha = 0.05$ (dashed). Endoscapes2023 has 20 disputed development cases; SAGES has 135.}\label{fig:method}
\end{figure*}

\paragraph{Videos, scores and monitoring} A case $i$ is the video before the first clip is applied, sampled at 1 frame per second: $x_{i,1},\dots,x_{i,T}$, with $T = 90$ in SAGES. A causal scorer produces a per-second score $z_{i,t}$. To ignore single-frame spikes, the monitor uses the minimum over the last $K = 3$ seconds, $w_{i,t} = \min_{t-K < u \le t} z_{i,u}$, and confirms CVS at the first second with $w_{i,t} \ge \tau$ (Fig.~\ref{fig:method}a). The case statistic
\begin{equation}
E_i = \max_{t \le T} w_{i,t}
\end{equation}
decides whether the monitor ever confirms within the window. In deployment this is a look-back evaluated when the clip applier is about to be used, as in a surgical time-out: has CVS been confirmed during the preceding $T$ seconds?

\paragraph{Reference standard and groups} Each video is rated by a panel of $m = 3$ experts, who mark each criterion at frames 5\,s apart and at the video level. Following the challenge, a video is \emph{sufficient} ($Y_i = 1$) if each criterion is marked by a majority of the panel at the video level, and \emph{insufficient} ($Y_i = 0$) otherwise. Let $V_i$ be the number of raters who marked all three criteria. Insufficient videos with $V_i = 1$ are \emph{disputed}; insufficient videos with $V_i = 0$ are \emph{clear}; $V_i \ge 2$ implies $Y_i = 1$. For a sufficient video, $t^f_i$ is the first labelled frame at which each criterion is marked by a majority of the panel. A confirmation before $t^f_i$ is \emph{premature}: it relies on evidence the raters did not yet see. With $E^{\pre}_i = \max_{t < t^f_i} w_{i,t}$, a confirmation is premature if $E^{\pre}_i \ge \tau$. Sufficient videos without a labelled majority-CVS frame have $t^f_i$ undefined and cannot be premature by this definition; we therefore also report premature confirmation among the videos with an observed onset. Vendor and country groups are defined by recorded metadata.

\paragraph{Risks} For a threshold $\tau$ and a group $g$ of insufficient videos (all insufficient videos, the disputed ones, the clear ones, or the insufficient videos of a vendor or country), the false-confirmation rate and the premature-confirmation rate are
\begin{align}
\FR_{g}(\tau) &= \P(E \ge \tau \mid Y = 0,\, i \in g),\\
\mathrm{PR}(\tau) &= \P(E^{\pre} \ge \tau \mid Y = 1).
\end{align}
We write $\FR_{\all}$, $\FR_{\dis}$ and $\FR_{\clr}$ for the first three groups. The benefit is \emph{timely sensitivity}: the probability that a sufficient video is confirmed, and not prematurely. Standard practice controls only $\FR_{\all}$, and only in expectation over the calibration set.

\section{Theory}\label{sec:theory}

\subsection{Two theorems on accuracy and disputed-case safety}

\begin{assumption}[panel sampling]\label{ass:panel}
For a video $X$, let $Q \in \{0,\dots,J\}$ be the number of experts in a pool of $J$ who would judge CVS achieved, and let $Q = q(X)$ be determined by the video. The panel is an exchangeable draw of $m$ of the $J$ experts, independent of everything else given $Q$. Then $V \mid X \sim \mathrm{Hypergeometric}(J, Q, m)$, and $V$ is conditionally independent of $X$ given $Q$. With $h = \lfloor m/2 \rfloor$, the vote-count reference calls a video sufficient if $V > h$, insufficient if $V \le h$, and disputed if $V = h$.
\end{assumption}

The assumption idealizes the SAGES design, in which three of twenty qualified experts were allocated to each video by an automated system \citep{alapatt2026sages} (Fig.~\ref{fig:method}c). Two departures matter. The allocation is not described as random. And the SAGES reference uses per-criterion majorities rather than the vote count: 11 of the 146 training videos and 7 of the 46 test videos with $V = 1$ are sufficient by the SAGES reference and so are not disputed. We report results under both references (Section~\ref{sec:res-sens}). The condition $Q = q(X)$ says that the video determines how the expert population would judge it; it is needed only so that the ideal rule below is itself a video-based rule.

Write $\FR_{\dis}(\phi) = \P(\phi = 1 \mid V = h)$ and $\Sens(\phi) = \P(\phi = 1 \mid V > h)$ for a confirmation rule $\phi$ that may use the video, possibly with randomization. Under Assumption~\ref{ass:panel}, any rule $\phi(X)$ has the same operating characteristics as the rule $\psi(Q) = \E[\phi(X) \mid Q]$, which uses only $Q$. It therefore suffices to compare rules that depend on $Q$. A rule that thresholds $Q$ is an \emph{ideal recognizer}: it knows how the expert population would judge the video.

\begin{theorem}[the ideal recognizer is the least safe]\label{thm:paradox}
Under Assumption~\ref{ass:panel}, fix $\alpha \in (0,1)$. Let $\phi^\star = \1\{Q > q^\star\} + \gamma\,\1\{Q = q^\star\}$ be the threshold rule on $Q$ with $\P(\phi^\star = 1 \mid V \le h) = \alpha$. Among all video-based rules with $\P(\phi = 1 \mid V \le h) = \alpha$:
\begin{enumerate}
\item[(i)] $\phi^\star$ maximizes $\Sens(\phi)$;
\item[(ii)] $\phi^\star$ maximizes $\FR_{\dis}(\phi)$;
\item[(iii)] $\FR_{\dis}(\phi^\star) \ge \alpha \ge \P(\phi^\star = 1 \mid V < h)$.
\end{enumerate}
\end{theorem}

The proof (\ref{app:proofs}) rests on the monotone likelihood ratio of the hypergeometric family in $Q$ \citep{karlin1956theory}. Both $\P(V > h \mid Q)/\P(V \le h \mid Q)$ and $\P(V = h \mid Q)/\P(V \le h \mid Q)$ are non-decreasing in $Q$ (Fig.~\ref{fig:method}e). By the Neyman--Pearson lemma \citep{neyman1933problem,lehmann2022testing}, the most powerful test of ``sufficient against insufficient'' and that of ``disputed against insufficient'', at the same level, are therefore the same threshold on $Q$. Theorem~\ref{thm:paradox} is extremal. The next result shows that the trend holds along a whole family of recognizers of increasing accuracy.

\begin{theorem}[more accurate, less safe]\label{thm:monotone}
Under Assumption~\ref{ass:panel}, let a recognizer's case score be $S_\sigma = \eta(Q) + \sigma Z$, where $\eta$ is strictly increasing, $Z \sim \mathcal{N}(0,1)$ is independent of the video and the panel, and $\sigma > 0$. Let the rule confirm when $S_\sigma$ exceeds the threshold at which $\P(\text{confirm} \mid V \le h) = \alpha$. As $\sigma$ decreases:
\begin{enumerate}
\item[(i)] the receiver operating characteristic (ROC) curve of $S_\sigma$ for sufficient against insufficient videos does not fall at any point, so the area under it (case AUROC) does not decrease;
\item[(ii)] sensitivity does not decrease;
\item[(iii)] disputed false confirmation does not decrease;
\item[(iv)] clear false confirmation does not increase.
\end{enumerate}
Disputed false confirmation tends to $\alpha$ as $\sigma \to \infty$, and to $\FR_{\dis}(\phi^\star)$ of Theorem~\ref{thm:paradox} as $\sigma \to 0$.
\end{theorem}

The proof composes two monotone likelihood ratios. The score is monotone in the latent share, and the latent share is monotone in the votes, so the score is monotone in the votes. By Karlin's composition theorem for totally positive kernels \citep{karlin1968total}, thresholding $S_\sigma$ is then the most powerful test of both hypotheses based on $S_\sigma$. A noisier score is a garbling of a less noisy one, and garbling cannot raise the power of a most powerful test \citep{blackwell1953equivalent}. Gaussian noise is a sufficient condition, not a necessary one: the proof uses only that the noise density is totally positive of order 2 and that noisier scores are garblings of less noisy ones.

\begin{corollary}[conservation of false confirmation]\label{cor:conservation}
For any rule, $\FR_{\dis} = \bigl(\FR_{\all} - (1-\pi)\,\FR_{\clr}\bigr)/\pi$, where $\pi$ is the disputed share of insufficient cases. At a fixed overall rate, every reduction of clear false confirmation raises disputed false confirmation by $(1-\pi)/\pi$ times as much, and $\FR_{\dis} \le \FR_{\all}/\pi$.
\end{corollary}

In SAGES training, $\pi = 135/495 = 0.27$. At $\FR_{\all} = 0.10$, halving clear false confirmation from 0.06 to 0.03 raises disputed false confirmation from 0.21 to 0.29.

\begin{remark}[heterogeneous raters]\label{rem:rasch}
Exchangeability can be replaced by a Rasch model \citep{rasch1960probabilistic}. Suppose rater $e$ marks CVS with probability $\Lambda(\theta - b_e)$, independently given the latent position $\theta$ of the video, where $\Lambda$ is the logistic function and $b_e$ the rater's strictness. For any \emph{fixed} panel $\mathcal{A}$, $V$ given $\theta$ is then an exponential family in $\theta$ with sufficient statistic $V$, so Theorems~\ref{thm:paradox} and~\ref{thm:monotone} hold within the panel with $Q$ replaced by $\theta$. This covers the same annotators labelling every case, as in Endoscapes2023. It does not cover random panels drawn from raters of different strictness.
\end{remark}

\subsection{What ``safe'' means here}\label{sec:target}

\paragraph{The target} The outcome that ultimately matters, bile duct injury, is rare and is recorded in no CVS dataset, so no study of a CVS check can measure it directly. We therefore use safety in an operational sense: a confirmation is false when the reference of an independent expert panel judges CVS not achieved, that is, when at least one criterion is not marked by a majority of the panel (Section~\ref{sec:problem}). This is how the SAGES and Endoscapes2023 videos are labelled \citep{alapatt2026sages,murali2023endoscapes}, it mirrors expert video review of CVS \citep{nijssen2015complications}, and it is the standard a confirming check must meet, because the check exists to catch videos in which the surgeon believes CVS is achieved and experts would not agree. The vote-count reference of the theory (Assumption~\ref{ass:panel}) differs from it for a few videos, and we report results under both (Section~\ref{sec:res-sens}). Our claims concern this target, not injuries.

\paragraph{Why disputed videos are constrained separately} Disagreement does not by itself make the anatomy of a video more dangerous, and we do not assume that it does. Three facts make a separate constraint necessary. First, a disputed video is insufficient by the reference, so confirming it is an error by the same standard as any other false confirmation, not a judgement call that the reference leaves open. Second, the average limit does not hold for these videos: the declared rate of 0.10 became 0.21 for disputed videos on the pre-specified test (Section~\ref{sec:res-test}), and Theorems~\ref{thm:paradox}--\ref{thm:monotone} show that this concentration is systematic and grows as recognition improves. The error rate a user is told is therefore predictably wrong for a large, identifiable class of videos. Third, these errors are not benign: every disputed video that standard calibration falsely confirmed on test lacked C1 or C3 by majority (post hoc; Section~\ref{sec:res-test}), and C3, two and only two structures, is the criterion most directly tied to misidentification.

\paragraph{Agreement with experts or with a noisy panel} A reader may object that the paradox merely penalizes recognizers for agreeing with the expert population rather than with a noisy three-rater panel. Three observations answer this.

First, the risk is measured against an independent panel. The panel that labels a video is a draw independent of the recognizer, so our false-confirmation rates are the probabilities that an \emph{independent} panel's majority rejects a confirmation: the standard by which the confirmation would be audited. A confirmation that an independent panel would overturn is unsafe in the sense that matters at the moment of clipping.

Second, the paradox itself is identified from the votes. Over all distributions of $Q$ that reproduce the observed training vote counts exactly ($J = 20$), the ideal recognizer's disputed false confirmation at marginal level 0.1 lies between 0.17 and 0.31 (linear programming; \ref{app:proofs}). Its lower bound exceeds $\alpha$ for every distribution of $Q$ that reproduces the observed vote shares (these bounds treat the shares as exact and ignore their sampling error).

Third, how many disputed confirmations are also wrong by the expert population is not identified, and we quote it only under the fitted model. Under the beta-binomial model (Section~\ref{sec:res-paradox}), 77\,\% of disputed videos have a latent expert share below one half; without the model, this share can be anywhere between 0.38 and 1.00. For a model recognizer about as accurate as our scorers (case AUROC 0.84), 56\,\% of the disputed videos that standard calibration confirms are insufficient by the population majority, and a fresh panel would reject 57\,\% of them. For the ideal recognizer these shares are 23\,\% and 39\,\%: the confirmed disputed videos shift towards genuinely contested ones. In the model, the paradox at realistic accuracy is therefore mostly a safety failure, and near the ideal it becomes a disagreement failure, in which confirmations override a substantial fraction of experts. Neither is acceptable for a confirming check.

\subsection{The price of protection}\label{sec:price}

\begin{proposition}[price of protection]\label{prop:price}
Let $W$ be the information a rule uses. Any rule $\phi(W)$ with disputed false confirmation at most $\alpha$ has sensitivity at most $\beta_W(\alpha)$, the power at level $\alpha$ of the most powerful test of ``sufficient'' against ``disputed'' based on $W$. A threshold rule on a scalar score is bounded by that score's ROC true-positive rate at disputed false-positive rate $\alpha$. Under Assumption~\ref{ass:panel}, $\beta_X(\alpha) \le \beta_Q(\alpha)$ for the video $X$.
\end{proposition}

Protecting disputed videos therefore costs sensitivity according to how poorly the score separates them from sufficient videos. The cost falls as recognition improves.

\begin{corollary}[protection turns accuracy into sensitivity]\label{cor:gain}
In the family of Theorem~\ref{thm:monotone}, let $\phi_\sigma$ confirm when $S_\sigma$ exceeds the threshold at which its disputed false confirmation equals $\alpha$. Then $\phi_\sigma$ attains the bound $\beta_{S_\sigma}(\alpha)$ of Proposition~\ref{prop:price}, its overall false confirmation is at most $\alpha$, and its sensitivity does not decrease as $\sigma$ decreases.
\end{corollary}

Within this model, lower recognition noise is spent on disputed errors under an average guarantee (Theorem~\ref{thm:monotone}) and on sensitivity under a disputed guarantee. Certification therefore does more than raise the threshold: it fixes where the error budget may go. The corollary is a population statement about sensitivity at one disputed limit, for recognizers that are noisier versions of one another. It does not say that every more accurate recognizer gains timely sensitivity under the full certified family, which also depends on premature confirmation, on the other groups and on the margin that a finite calibration set requires; on test, the less accurate temporal scorer had the higher certified timely sensitivity (Section~\ref{sec:res-test}). Our scorers are far from this ceiling: the frame scorer separates sufficient from disputed videos with case AUROC 0.69, against 0.87 for the ideal recognizer of the fitted model. No rule can beat the separability of its information, which explains why every attempt to recover sensitivity by reshaping the rule failed (\ref{app:negative}). The bound $\beta_Q(\alpha)$, the protected sensitivity of the ideal recognizer, is a ceiling for every video-based recognizer, but it is not identified from three votes per case, which fix only the first three moments of the distribution of $Q$. Over all distributions of $Q$ that reproduce the observed vote counts exactly, the ceiling at $\alpha = 0.1$ ranges from 0.51 to 0.95 in training and from 0.39 to 0.81 in test (\ref{app:proofs}). Under the fitted beta-binomial model with $J = 20$ and a randomized threshold it is 0.68 (bootstrap 95\,\% confidence interval, CI, 0.63--0.72) in training and 0.66 (0.59--0.72) in test; the pre-specified version of this computation, with a continuous latent share and a deterministic threshold, gives 0.65 and 0.61. Larger panels would identify it.

\subsection{Monitoring and calibration size}

\begin{proposition}[monitoring]\label{prop:monitor}
For any score stream and threshold, the event ``the monitor confirms at some second in the window'' equals $\{E \ge \tau\}$, and the event ``it confirms before $t^f$'' equals $\{E^{\pre} \ge \tau\}$. A guarantee on these events is therefore a guarantee for a monitor that checks every second of the window, without correction for repeated looks. It does not extend to windows longer than those calibrated.
\end{proposition}

\begin{lemma}[minimum group size]\label{lem:nmin}
With binomial $p$-values, a group with $n_g$ calibration cases can be certified at level $(\alpha, \delta)$ only if $(1-\alpha)^{n_g} \le \delta$, that is, $n_g \ge \lceil \log \delta / \log(1-\alpha) \rceil$. For $\delta = 0.1$ this is 22 cases at $\alpha = 0.1$, 45 at $\alpha = 0.05$ and 230 at $\alpha = 0.01$. If any certified group is smaller, the certified rule never confirms.
\end{lemma}

The 20 disputed development cases of Endoscapes2023 fall below this minimum (Fig.~\ref{fig:method}f). SAGES, with 135 disputed training videos, supports $\alpha = 0.1$ and $0.05$ but not $0.01$.

\section{Method}\label{sec:method}

\subsection{Certified confirmation}

Let $\mathcal{D}$ be a calibration set of videos with case statistics $E_i$ and $E_i^{\pre}$, reference labels and group memberships. Each certified risk $g \in \mathcal{G}$ is a pair $(T_g, \mathcal{D}_g)$: a statistic and the calibration videos it concerns. We certify the false-confirmation risks of all insufficient videos, of the disputed videos, and of the insufficient videos of each vendor and country group, all with statistic $T_g = E$, and the premature-confirmation risk of the sufficient videos, with statistic $T_g = E^{\pre}$. For a candidate threshold $\tau$, let $N_g(\tau) = \#\{i \in \mathcal{D}_g : T_{g,i} \ge \tau\}$ and $n_g = |\mathcal{D}_g|$. Then
\begin{equation}
p_g(\tau) = \P\bigl(\mathrm{Bin}(n_g, \alpha) \le N_g(\tau)\bigr)
\end{equation}
is a valid $p$-value for the null hypothesis $H_g(\tau)\!: R_g(\tau) > \alpha$, where $R_g(\tau) = \P(T_g \ge \tau \mid i \in g)$ is the risk. The null that some certified risk exceeds $\alpha$ is a union of these nulls, so $p(\tau) = \max_g p_g(\tau)$ is valid for it (intersection--union). Algorithm~\ref{alg:cert} scans the candidates from strict to permissive and returns the most permissive certified threshold (Fig.~\ref{fig:method}b).

\begin{algorithm}[!tb]
\caption{Certified confirmation (calibration)}\label{alg:cert}
\begin{algorithmic}[1]
\Require calibration statistics $\{E_i, E^{\pre}_i\}$, certified risks $\mathcal{G}$, levels $\alpha, \delta$
\State $\Lambda_\tau \gets$ distinct finite values of $E$ and $E^{\pre}$ in $\mathcal{D}$, in decreasing order
\State $\hat\tau \gets +\infty$
\For{$\tau \in \Lambda_\tau$}
  \State $p(\tau) \gets \max_{g \in \mathcal{G}} \P(\mathrm{Bin}(n_g, \alpha) \le N_g(\tau))$
  \If{$p(\tau) > \delta$} \textbf{break} \EndIf
  \State $\hat\tau \gets \tau$
\EndFor
\State \Return $\hat\tau$ \Comment{use: confirm when $w_t \ge \hat\tau$}
\end{algorithmic}
\end{algorithm}

\begin{theorem}[validity]\label{thm:valid}
Suppose the scorer does not depend on the calibration labels, and the calibration videos and a future video are independent draws from the same distribution of $(E, E^{\pre}, Y, V, \text{group memberships})$. Then with probability at least $1-\delta$ over the calibration set, every certified risk at $\hat\tau$ is at most $\alpha$.
\end{theorem}

The argument is that of risk-controlling prediction and Learn-then-Test \citep{bates2021distribution,angelopoulos2025learn}. Each certified risk is monotone in $\tau$, so validity holds for any certified threshold, even though the candidates are taken from the calibration data (\ref{app:proofs}). Because $p(\tau)$ is itself monotone, the scan is a convenience rather than a source of multiplicity. The rule uses no group membership at use: it is one threshold for all videos. The guarantee for disputed videos comes from calibration labels alone. Because the disputed and premature risks are conditional on the reference class, a shift that changes only the proportions of clear, disputed and sufficient videos leaves them unchanged, although it changes the overall and group risks, which mix clear and disputed videos.

\subsection{Relation to existing calibration rules}\label{sec:relation}

Each per-group test is the binomial bound behind training-conditional (PAC) conformal calibration \citep{vovk2012conditional,park2020pac} and Neyman--Pearson umbrella classification \citep{tong2018neyman}. Taking the maximum $p$-value across groups yields the most conservative of the per-group PAC thresholds, so at the level of the algorithm certified confirmation is group-wise PAC calibration. What it adds is the choice of what to certify, justified by Theorems~\ref{thm:paradox}--\ref{thm:monotone}: a group that is unobservable at use, a timing risk on a monitored stream, and vendor and country groups. Three comparators follow from this view. \emph{PAC marginal} calibration certifies only the overall risk with high probability; it tests whether the failure of standard calibration comes from controlling the average or from controlling it only in expectation. \emph{Group-wise conformal} calibration takes the maximum of the per-group conformal thresholds; it protects every group, but only in expectation over the calibration set. \emph{Weighted conformal risk control} \citep{angelopoulos2024conformal} up-weights disputed false confirmations in a single expected loss; it shifts the average towards disputed videos but still controls an average. Multi-group and fair risk control \citep{zhang2024fair,jung2023batch} need group membership as a function of the input when the rule is applied, which disputed status is not.

\subsection{Scorers}\label{sec:scorers}

The method applies to any scorer. We study scorers built on DINOv3 \citep{simeoni2025dinov3}.

\paragraph{Frame scorer (primary)} A DINOv3 ViT-B/16 backbone (LVD-1689M weights) is fine-tuned in its last 8 of 12 blocks and final norm, on 1-fps frames of $448 \times 256$ pixels. A linear head on the class token and the mean patch token (1,536 dimensions) outputs C1--C3 and their conjunction. Soft targets are the fraction of raters marking each criterion at the nearest labelled frame (at most 2.5\,s away), with binary cross-entropy weighted for positives by $\sqrt{\min(50, \max(1, (1-p)/p))}$, where $p$ is the training prevalence. Optimization uses AdamW \citep{loshchilov2019decoupled} (learning rates $2\times10^{-5}$ for the backbone and $10^{-3}$ for the head, weight decay 0.05), 200 warm-up steps with linear decay, and 3,000 steps of batch 32 in bfloat16 with gradient clipping at 1.0. Augmentation is random cropping to 85--100\,\%, flips, and brightness and contrast jitter; the checkpoint is selected by mAP on a separate fold. The per-second score is $z_t = \mathrm{logit}(p_1 p_2 p_3)$, the logit of the product of the three criterion probabilities.

\paragraph{Temporal scorer (secondary, exploratory)} A causal network reads the frame scorer's per-second features and logits, standardized, through LayerNorm, dropout 0.3, a linear layer to 128 units with GELU, and a gated recurrent unit \citep{cho2014learning}. Each second's criterion logits come from the input embedding and the recurrent state. It is trained only on the current view at labelled frames, never on video-level targets, which would teach it to anticipate CVS. Training uses AdamW (learning rate $10^{-3}$, weight decay $10^{-2}$) for up to 60 epochs with early stopping (patience 10) on an inner fold's AUROC of the monitored statistic for sufficient against disputed videos; three seeds are averaged. Its score is the logit of the product of its criterion probabilities.

\paragraph{Further development scorers} A frozen scorer uses frozen DINOv3-B features with a two-layer perceptron (256 hidden units; three seeds). Eight further temporal heads, trained with frame-level, ordinal or ranking losses (\ref{app:negative}), provide additional development scorers of varying accuracy, eleven in total.

\section{Experimental design}\label{sec:design}

\subsection{Data}

\begin{table*}[!tb]
\centering
\caption{Datasets. Disputed means insufficient by the reference standard, but: in SAGES, exactly one of three raters marked all criteria at the video level; in Endoscapes2023, at least one of three fixed annotators did so at some keyframe; in Cholec80-CVS (criteria scored 0--2), the maximum total score reached at least 3 of 6. Endoscapes2023 and Cholec80-CVS: development cases only.}\label{tab:data}
\footnotesize
\begin{tabular}{lrrrrl}
\toprule
Dataset & Videos & Sufficient & Disputed & Clear & Use \\
\midrule
SAGES CVS Challenge 2024, training & 700 & 205 & 135 & 360 & development (5-fold cross-fitting) \\
SAGES CVS Challenge 2024, test & 300 & 62 & 39 & 199 & single pre-specified evaluation \\
Endoscapes2023, development & 155 & 46 & 20 & 89 & replication of the paradox (hypothesis-generating) \\
Cholec80-CVS, development & 56 & 14 & 19 & 23 & replication of the paradox (hypothesis-generating) \\
\bottomrule
\end{tabular}
\end{table*}

\paragraph{SAGES CVS Challenge 2024} The dataset \citep{alapatt2026sages} contains 1,000 videos of 90\,s before the first clip from 54 institutions in 24 countries (CC BY-NC 4.0). The official challenge split gives 700 training videos (23 countries) and 300 test videos (18 countries). Every 5\,s (18 frames per video), three of 20 qualified raters marked each criterion; they also gave video-level labels with a confidence. Rater identities are not released. Votes for full CVS ($V = 0/1/2/3$) were 360/146/90/104 in training and 199/46/31/24 in test (Table~\ref{tab:data}). Device vendor is missing for 156 training and 114 test videos. Of the sufficient videos, 151 of 205 in training and 38 of 62 in test have a labelled frame with majority CVS; the others cannot be premature.

\paragraph{Label shift between training and test} The vote distributions of the two sets differ ($\chi^2_3 = 20.5$, $p = 1.3\times10^{-4}$). Sufficient videos made up 29\,\% of training and 21\,\% of test videos (Fisher's exact $p = 0.005$), and disputed videos 27\,\% and 16\,\% of insufficient ones ($p = 0.001$). The test set therefore held 39 disputed videos rather than the 58 expected from the training share of disputed videos (135 of 700), which lowered the power of the primary endpoint. The shift also departs from the exchangeability that Theorem~\ref{thm:valid} assumes. A shift in class proportions alone would leave the disputed and premature risks unchanged (Section~\ref{sec:method}), but the test is also a check under a shift that we did not anticipate.

\paragraph{Replication datasets} The Endoscapes2023 development cases \citep{murali2023endoscapes} and the Cholec80-CVS development cases \citep{rios2023cholec80cvs} were scored by seven recognizers trained by cross-fitting before this study. Their sealed test sets were not used. In Endoscapes2023, 18 of the 20 disputed cases come from one annotator who alone marked full CVS, so disputed status there reflects one lenient rater (Remark~\ref{rem:rasch}). We treat both replications as hypothesis-generating.

\subsection{Development by cross-fitting}

The training videos were split into five folds, stratified by sufficiency and country. For fold $f$, the frame-scorer backbone was trained on three folds, selected on fold $f{+}1$ and applied to fold $f$, which it never saw. The temporal head for fold $f$ was trained on the same three folds and early-stopped on fold $f{+}1$, using those folds' out-of-fold frame features. Those features come from backbones whose own training folds included fold $f$, so the temporal scores of fold $f$ depend indirectly on fold-$f$ labels. This dependence, and its selection on development data, are why the temporal scorer is secondary and exploratory. Rules were compared by calibrating on four folds of out-of-fold scores and evaluating on the fifth. Out-of-fold frame mAP was 0.43 for the frozen scorer and 0.57 for the frame scorer, below the best challenge entries on the test set (0.69) and close to the challenge baselines (0.59) \citep{alapatt2026sages}. Our recognizers are not the contribution, and Theorem~\ref{thm:monotone} suggests that stronger recognizers of the same kind would fare worse under standard calibration.

\subsection{Pre-specification and the single test run}\label{sec:prereg}

\paragraph{Evolution of the protocol} The protocol evolved on training data only, with every change logged. Before any result for the frame scorer was seen, a boundary-aware temporal model was fixed as the proposed scorer, with a decision rule for carrying it to the test. That model missed its rule (\ref{app:negative}), so the frame scorer became primary; the temporal scorer was declared secondary and flagged as selected on development data.

\paragraph{The lock} We calibrated all thresholds on the 700 out-of-fold scores ($\alpha = \delta = 0.1$). We then hash-locked (SHA-256) the protocol, endpoints and power calculation together with the code, models, out-of-fold scores and thresholds: 34 files. Before the lock, the test label files had not been downloaded. The lock is internal, with no third-party timestamp; the lock file and the test-access log will be made public with the code upon publication.

\paragraph{Scoring the test set} Each test video was scored by one cross-fitted pipeline, $f = \mathrm{SHA256}(\text{video id}) \bmod 5$, so the deployed rule is a randomized ensemble. Averaging the five pipelines would have changed the score distribution and voided the calibration. Before the lock, the test-time code reproduced the out-of-fold scores of training videos exactly (maximum absolute difference 0). The scores were saved and their hash logged before the label files were downloaded. The test set was evaluated once, with no deviation from the protocol. Every later analysis of test labels is post hoc and recorded in the test-access log (Table~\ref{tab:status}).

\paragraph{What the guarantee covers} Theorem~\ref{thm:valid} requires a scorer independent of the calibration labels and independent, identically distributed cases. The deployed system departs from this in three ways. Its calibration scores come from five cross-fitted pipelines, the usual approximation \citep{vovk2015cross,barber2021predictive}, and test videos are allocated to pipelines at random rather than in fixed proportions. Videos from one centre, or rated by shared raters, are not independent, which makes binomial tests anti-conservative. And the label shift above departs from identical distribution. No finite-sample guarantee therefore covers the system exactly as deployed. We test the guarantee empirically instead: on vendor and country groups left out of calibration, on the held-out test set with country-clustered intervals, and with a split calibration in which each pipeline is calibrated only on videos it never saw (Section~\ref{sec:res-valid}).

\subsection{Comparators, endpoints and statistics}

\paragraph{Comparators} Standard practice is conformal marginal calibration: the $\lceil (n_0+1)(1-\alpha)\rceil$-th smallest $E$ among the $n_0$ insufficient calibration videos, with confirmation if $E$ is strictly greater, which guarantees $\E[\FR_{\all}] \le \alpha$ \citep{angelopoulos2023gentle}. On development we also compare a fixed probability threshold of 0.5 ($\tau = 0$), Youden's threshold, PAC marginal calibration, group-wise conformal calibration and weighted conformal risk control (Section~\ref{sec:relation}). The certified rule certifies overall, disputed and premature risk and the six vendor and country groups with at least 60 insufficient training videos (vendors 0, 2, 4; countries 6, 0, 2).

\paragraph{Pre-specified endpoints} All endpoints concern the 300 test videos and $\alpha = 0.1$. E1 (primary, frame scorer) has two parts. E1a: under standard calibration, disputed false confirmation exceeds $\alpha$ (one-sided exact binomial test of $H_0\!: p \le 0.1$ at level 0.05; power 0.68 for the expected 58 disputed videos, computed before the lock). E1b: under the certified rule, the overall, disputed and premature risks show no evidence of exceeding $\alpha$ (the same test with Holm correction \citep{holm1979simple}). E2: standard premature confirmation exceeds $\alpha$ (power 0.50). E3 (descriptive): vendor and country groups. E4 (secondary): timely sensitivity under the certified rule, temporal against frame scorer, by a one-sided exact McNemar test \citep{mcnemar1947note}. E5: criterion guidance (\ref{app:guidance}). E6 (descriptive): theory checks.

\paragraph{Statistics} Intervals for rates are exact two-sided Clopper--Pearson 95\,\% CIs \citep{clopper1934use}; upper bounds (UB) are one-sided 95\,\% bounds. Paired development differences use a bootstrap over videos with thresholds held fixed, which is optimistic because it ignores recalibration and selection. Rank correlations across scorers use a bootstrap over cases with recalibration inside each of 1,000 resamples. On the test set, country-clustered bootstrap intervals (2,000 resamples of countries) account for dependence within centres. To illustrate the difference between protection in expectation and with high probability, each rule is also calibrated on 1,000 bootstrap samples of the 700 development videos and its risks are evaluated exactly on the 700, treated as the population; this resampling reuses the development videos, so it illustrates the two kinds of control rather than testing validity on new data.

\begin{table*}[!tb]
\centering
\caption{Status of every analysis. Only the first row was fixed before the lock. Development analyses after the lock use no test data; all except the Endoscapes2023 and Cholec80-CVS replication were run after unblinding, so their choice was made knowing the test results. Every post hoc analysis of test labels is recorded in the test-access log, which will be made public with the code.}\label{tab:status}
\footnotesize
\begin{tabular}{L{0.21\textwidth}L{0.55\textwidth}L{0.17\textwidth}}
\toprule
Status & Analyses & Where \\
\midrule
Pre-specified, hash-locked, single test run & Scorers, thresholds and endpoints E1--E6; development results of the standard and certified rules for the frame and temporal scorers; method variants that failed & Tables~\ref{tab:dev}, \ref{tab:endpoints}; Figs.~\ref{fig:overview}c (top), d, \ref{fig:method}a,b, \ref{fig:test}; \ref{app:negative} \\
Development, after the lock & Endoscapes2023 and Cholec80-CVS replication; frozen-scorer results; further comparators (fixed probability, Youden, PAC marginal, group-wise conformal, weighted risk control); the eleven-scorer association; left-out vendors and countries; resampling illustration; ablation; other levels $\alpha$; positive predictive value and false-discovery-rate certification; model-based safety quantities and identification bounds & Tables~\ref{tab:dev}, \ref{tab:validity}, \ref{tab:ablation}--\ref{tab:replication}; Figs.~\ref{fig:overview}c (bottom), \ref{fig:method}d, \ref{fig:paradox}a,c, \ref{fig:price}b \\
Post hoc on test labels (logged) & Group-wise conformal rule; model curve refitted to test votes; vote-count reference; premature confirmation among videos with an observed onset; positive predictive value; country-clustered intervals; split calibration; missing criteria; comparison at matched disputed rate; label-shift tests & Sections~\ref{sec:res-valid}--\ref{sec:res-sens}; Table~\ref{tab:readout}; Fig.~\ref{fig:price}a \\
\bottomrule
\end{tabular}
\end{table*}

\paragraph{Software} Analyses were implemented in Python (PyTorch, NumPy, SciPy); parts of the code were written with the help of an AI coding assistant (Claude Code, Anthropic). Every reported number is produced by scripts from logged outputs and cross-checked against a numbers ledger, and every reference was checked against Crossref, PubMed or arXiv records. Figures were drawn by plotting scripts without generative image tools.

\section{Results}\label{sec:results}

\subsection{Development: average guarantees hide disputed videos}\label{sec:res-dev}

\begin{table*}[!tb]
\centering
\caption{SAGES development (700 training videos, 5-fold cross-fitting, $\alpha = \delta = 0.1$). Bold: disputed false confirmation above $\alpha$. Premature: among all sufficient videos, as certified. Worst group: highest false confirmation among the six certified vendor and country groups. Timely: confirmed, and not before the first majority-CVS frame. The frozen-scorer rows and all rules other than standard and certified were computed after unblinding (Table~\ref{tab:status}).}\label{tab:dev}
\footnotesize
\begin{tabular}{llcccccc}
\toprule
 & & \multicolumn{3}{c}{False confirmation} & & Worst & Timely \\
\cmidrule(lr){3-5}
Scorer & Rule & all & disputed & clear & Premature & group & sensitivity \\
\midrule
Frozen & standard (conformal) & 0.101 & \textbf{0.170} & 0.075 & 0.122 & 0.143 & 0.229 \\
       & certified & 0.028 & 0.059 & 0.017 & 0.024 & 0.048 & 0.112 \\
\midrule
Frame & probability $> 0.5$ & 0.057 & \textbf{0.148} & 0.022 & 0.093 & 0.059 & 0.254 \\
      & Youden & 0.319 & \textbf{0.563} & 0.228 & 0.429 & 0.419 & 0.351 \\
      & standard (conformal) & 0.099 & \textbf{0.207} & 0.058 & 0.166 & 0.114 & 0.332 \\
      & PAC marginal & 0.081 & \textbf{0.178} & 0.044 & 0.137 & 0.093 & 0.332 \\
      & group-wise conformal & 0.034 & \textbf{0.104} & 0.008 & 0.059 & 0.057 & 0.180 \\
      & certified & 0.016 & 0.052 & 0.003 & 0.039 & 0.025 & 0.107 \\
\midrule
Temporal & probability $> 0.5$ & 0.131 & \textbf{0.267} & 0.081 & 0.185 & 0.195 & 0.351 \\
         & standard (conformal) & 0.101 & \textbf{0.207} & 0.061 & 0.161 & 0.161 & 0.322 \\
         & PAC marginal & 0.083 & \textbf{0.170} & 0.050 & 0.137 & 0.144 & 0.312 \\
         & group-wise conformal & 0.046 & 0.081 & 0.033 & 0.088 & 0.076 & 0.234 \\
         & certified & 0.024 & 0.044 & 0.017 & 0.049 & 0.048 & 0.176 \\
\bottomrule
\end{tabular}
\end{table*}

Under standard calibration, overall false confirmation sat at the nominal level for every scorer (0.099--0.101; Table~\ref{tab:dev}), while disputed false confirmation was 0.17--0.21 and premature confirmation 0.12--0.17. Controlling the overall risk with high probability (PAC marginal) changed little (disputed 0.17--0.18): the failure comes from controlling an average, not from controlling it only in expectation. Thresholds used in practice also exceeded $\alpha$ among disputed videos: Youden's threshold confirmed 56\,\% of them, and a fixed probability of 0.5 gave 0.148 (frame) and 0.267 (temporal). Weighted conformal risk control, with disputed false confirmations weighted 5 or 11 times as heavily as clear ones, left disputed false confirmation at 0.16--0.19 for both scorers.

The certified rule kept every certified risk at or below 0.059 when pooled over held-out folds. Its timely sensitivity was 0.107 (frame) and 0.176 (temporal; optimistic, Section~\ref{sec:design}). For the frame scorer, certifying the overall risk alone left timely sensitivity at 0.332, and adding the disputed risk reduced it to 0.107, the full cost (Table~\ref{tab:ablation}). The disputed risk binds for the frame scorer (Fig.~\ref{fig:method}b); for the temporal scorer, one country group binds.

\subsection{The paradox, predicted and observed}\label{sec:res-paradox}

\begin{figure*}[!t]
\centering
\includegraphics[width=\textwidth]{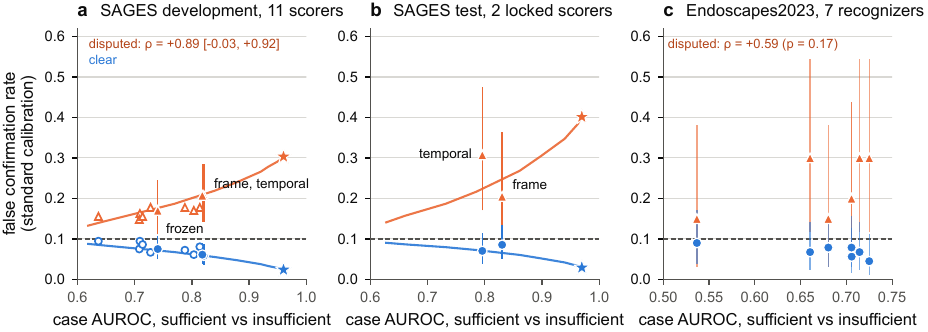}
\caption{The paradox, predicted and observed: false confirmation under standard calibration ($\alpha = 0.1$, dashed) for disputed (triangles, orange) and clear insufficient (circles, blue) videos, against case AUROC.
(a) SAGES development: 11 cross-fitted scorers; the frozen, frame and temporal scorers are filled, with exact 95\,\% CIs (vertical lines). Spearman $\rho$ for disputed videos with its bootstrap 95\,\% CI (recalibrated in each resample). Lines: prediction of the panel model fitted to training votes for recognizers that recover the latent share with decreasing noise (Theorem~\ref{thm:monotone}); stars: the model's ideal recognizer (Theorem~\ref{thm:paradox}). Lines and stars are model-based.
(b) SAGES test: the two locked scorers, with the model refitted to test votes (descriptive, added after the test run).
(c) Endoscapes2023 development cases: seven recognizers with exact 95\,\% CIs, against case AUROC (Table~\ref{tab:replication}; hypothesis-generating; 18 of 20 disputed cases come from one annotator).}\label{fig:paradox}
\end{figure*}

\paragraph{The model} We fitted a beta-binomial model, $Q/J \sim \mathrm{Beta}(a, b)$ with $J = 20$, by maximum likelihood to the training vote counts ($a = 0.35$, $b = 0.79$). It reproduces the vote distribution (fitted 362/136/102/101 against observed 360/146/90/104), and its likelihood ratios are monotone. A numerical check found no rule, among 20,000 random rules at the same level, exceeding the ideal recognizer's disputed false confirmation (0.303) or sensitivity (0.866). As case AUROC rises along the family of Theorem~\ref{thm:monotone}, disputed false confirmation increases from 0.13 at case AUROC 0.62 towards 0.30, and clear false confirmation falls towards 0.02 (Fig.~\ref{fig:paradox}a); as the noise grows without bound, both tend to $\alpha = 0.10$. These curves are model-based; the identification bounds of Section~\ref{sec:target} show which conclusions do not depend on the model.

\paragraph{SAGES development} The 11 development scorers fall along this curve without any fitting to it. Case AUROC was rank-correlated with disputed false confirmation ($\rho = +0.89$) and negatively with clear false confirmation ($\rho = -0.76$). The scorers share development data and are not independent: a bootstrap over videos with recalibration gave $\rho$ a 95\,\% CI of $-0.03$ to $0.92$, with $\rho \le 0$ in 3.4\,\% of resamples. The frozen-to-frame difference alone ($0.170 \to 0.207$) was not significant. The cross-scorer association is consistent with the theory but uncertain.

\paragraph{SAGES test} The test scorers lie within their CIs of the model curve refitted to test votes (Fig.~\ref{fig:paradox}b; descriptive, added after the test run). The two scorers are not ordered by the paradox: the less accurate temporal scorer confirmed more disputed videos (12/39) than the frame scorer (8/39). Theorem~\ref{thm:monotone} concerns recognizers that are garblings of one another, which two different scorers need not be.

\paragraph{Endoscapes2023 (hypothesis-generating)} Across seven recognizers (Table~\ref{tab:replication}), frame mAP was rank-correlated with disputed false confirmation ($\rho = +0.85$; bootstrap 95\,\% CI 0.06 to 0.95) and negatively with clear false confirmation ($\rho = -0.84$). Against case AUROC the association was weaker ($\rho = +0.59$, $p = 0.17$; Fig.~\ref{fig:paradox}c). Case AUROC between sufficient and clear cases ranged from 0.55 to 0.77, while that between sufficient and disputed cases stayed between 0.49 and 0.61: better recognizers separated the easy negatives and left the disputed boundary where it was. On Cholec80-CVS (56 cases, disputed by partial scores) the effect was not resolved ($\rho = +0.31$, $p = 0.50$).

\subsection{Pre-specified test}\label{sec:res-test}

\begin{figure*}[!t]
\centering
\includegraphics[width=\textwidth]{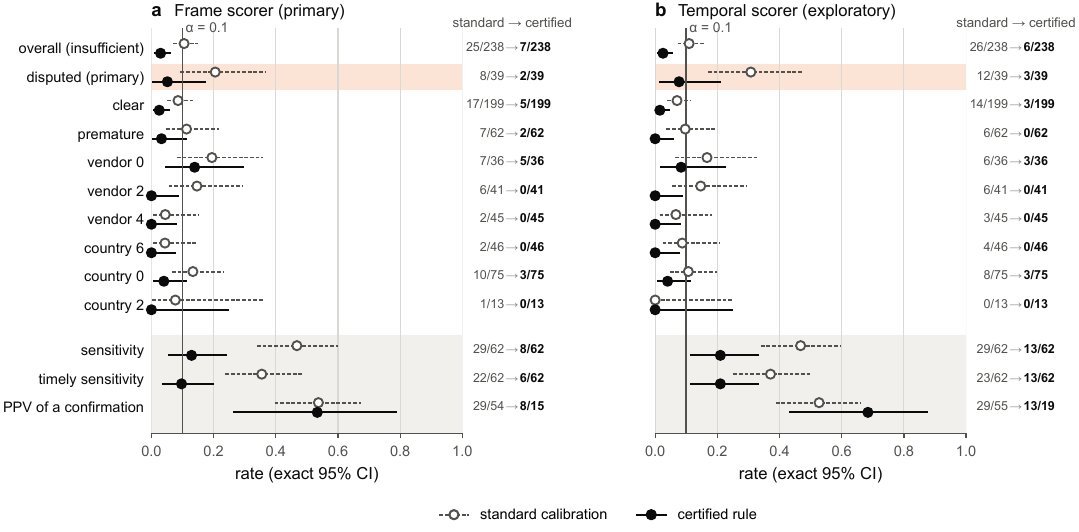}
\caption{Pre-specified SAGES test (300 videos; $\alpha = 0.1$). Every certified risk, plus clear videos (not certified), and on grey the benefits (sensitivity, timely sensitivity of E4 and the positive predictive value, PPV, of a confirmation), under standard calibration (open circles, dashed) and the certified rule (filled), with exact two-sided 95\,\% CIs and counts (standard $\rightarrow$ certified). Shaded orange: the primary risk. Vendor and country groups are those with at least 60 insufficient training videos. PPV is a post hoc read-out. (a) Frame scorer, primary. (b) Temporal scorer, secondary and exploratory.}\label{fig:test}
\end{figure*}

\begin{table*}[!tb]
\centering
\caption{Pre-specified endpoints on the SAGES test set (one run, no deviation). $p$: one-sided exact tests. UB: one-sided 95\,\% upper confidence bound.}\label{tab:endpoints}
\footnotesize
\begin{tabular}{L{0.30\textwidth}L{0.47\textwidth}L{0.15\textwidth}}
\toprule
Endpoint & Result & Outcome \\
\midrule
E1a: standard disputed false confirmation $> 0.1$ (frame) & 8/39 = 0.205 (95\,\% CI 0.093--0.365), $p = 0.037$ & met \\
E1b: certified overall, disputed and premature risk $\le 0.1$ & Holm $p = 1.00$; disputed 2/39 (UB 0.153), overall 7/238 (UB 0.055), premature 2/62 (UB 0.098) & no violation found \\
E2: standard premature confirmation $> 0.1$ (frame) & 7/62 = 0.113, $p = 0.43$ & inconclusive \\
E3: vendor and country groups & certified, frame: vendor 0 = 5/36 (UB 0.270); other groups at most 3 & reported \\
E4: temporal vs frame, timely (certified) & 13/62 vs 6/62, McNemar 8:1, $p = 0.020$ & met; exploratory (see text) \\
E5: guidance false alerts $\le 0.1$ & frame 0.06--0.09; temporal 0.06--0.12 & partly met \\
E6: theory checks & ceiling refitted to test votes 0.61 (training 0.65); certified sensitivity 0.13 (frame), 0.21 (temporal) vs 0.24, 0.21 attainable at the disputed limit & reported \\
\bottomrule
\end{tabular}
\end{table*}

The test set had 62 sufficient, 39 disputed and 199 clear insufficient videos (Fig.~\ref{fig:test}, Table~\ref{tab:endpoints}).

\paragraph{E1a: met} Under standard calibration the primary scorer confirmed 25 of 238 insufficient videos (0.105), close to $\alpha$, the level it controls on average. Among disputed videos it confirmed 8 of 39 (0.205; 95\,\% CI 0.093--0.365; $p = 0.037$), against 17 of 199 clear videos (0.085).

\paragraph{E1b: no evidence of violation} Under the certified rule, disputed false confirmation was 2/39 (0.051; one-sided 95\,\% upper bound 0.153). Overall false confirmation was 7/238 (0.029; upper bound 0.055), and premature confirmation 2/62 (upper bound 0.098). This is absence of evidence, and the test was not designed to establish compliance: against a true disputed rate of 0.20, Holm's first step had power 0.38.

\paragraph{E2: inconclusive} Standard premature confirmation was 7/62 (0.113; $p = 0.43$).

\paragraph{E3: vendor and country groups} Under the certified rule, five of six groups had at most 3 false confirmations each. Vendor 0 had 5/36 (0.139) with the frame scorer, a point estimate above $\alpha$ without evidence of violation ($p = 0.29$; upper bound 0.270).

\paragraph{E4 (secondary): met} The temporal scorer confirmed 13 of 62 sufficient videos on time under the certified rule, against 6 for the frame scorer (discordant 8 against 1; $p = 0.020$), and none prematurely. The two rules ran at different realized disputed rates (3/39 and 2/39). In a post hoc comparison at matched disputed false-positive rate, the temporal scorer's sensitivity, which includes premature confirmations, was not higher (Fig.~\ref{fig:price}a); timely sensitivity at a matched rate was not compared. Because the temporal scorer depends indirectly on development labels, a dependence we identified after unblinding, we regard E4 as exploratory.

\paragraph{Which criteria were missing (post hoc)} The 8 disputed videos falsely confirmed by the standard rule lacked C3 (five videos) or C1 (three). C3, ``two and only two structures'', is the criterion most directly tied to misidentification.

\subsection{Does the protection hold?}\label{sec:res-valid}

\begin{figure*}[!t]
\centering
\includegraphics[width=0.94\textwidth]{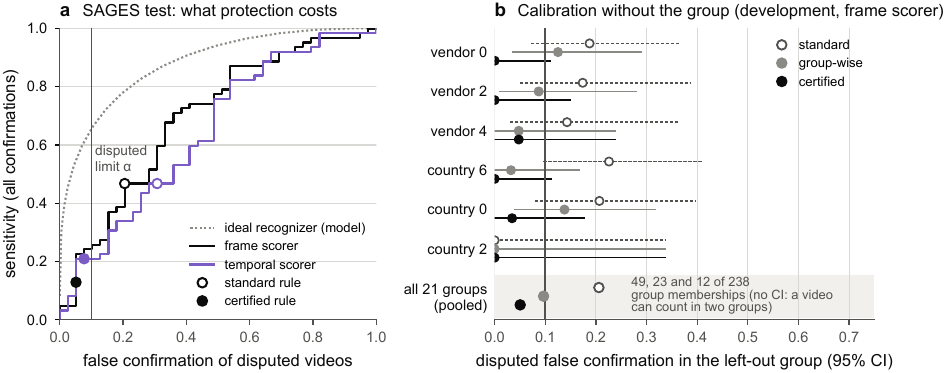}
\caption{What protection costs and whether it holds.
(a) SAGES test (post hoc): sensitivity against disputed false confirmation as the threshold varies, for both scorers and for the ideal recognizer of the panel model refitted to test votes (model-based). Open and filled circles: standard and certified operating points.
(b) Development, frame scorer: disputed false confirmation in a vendor or country group whose videos were all removed from calibration, under standard calibration (open), group-wise conformal calibration (grey) and the certified rule (filled; the left-out group is also removed from the certified family), with exact 95\,\% CIs. Bottom row: pooled over all 21 vendor and country groups that contain a disputed video (238 disputed group memberships, since each disputed video belongs to one country and, where recorded, one vendor); no interval is shown because memberships are not independent.}\label{fig:price}
\end{figure*}

\begin{table}[!tb]
\centering
\caption{Protection under left-out groups and under resampling (development). Left out: disputed false confirmations when each of the 21 vendor and country groups with a disputed video is removed from calibration in turn, pooled over 238 disputed group memberships of 135 disputed videos (not independent). Resampling: share of 1,000 bootstrap calibrations in which some certified risk exceeds $\alpha$ in the development population (an illustration of expectation against high-probability control), and mean timely sensitivity. --: not computed.}\label{tab:validity}
\footnotesize
\setlength{\tabcolsep}{3pt}
\begin{tabular}{llccc}
\toprule
 & & Left out & Resampling & \\
Scorer & Rule & disputed FC & any $> \alpha$ & Timely \\
\midrule
Frame & standard & 49/238 & 1.000 & 0.337 \\
      & PAC marginal & -- & 1.000 & 0.325 \\
      & group-wise & 23/238 & 0.318 & 0.178 \\
      & certified & 12/238 & 0.016 & 0.127 \\
\midrule
Temporal & standard & 51/238 & 1.000 & 0.329 \\
         & PAC marginal & -- & 1.000 & 0.318 \\
         & group-wise & 17/238 & 0.191 & 0.235 \\
         & certified & 10/238 & 0.002 & 0.173 \\
\bottomrule
\end{tabular}
\end{table}

The guarantee of Theorem~\ref{thm:valid} does not cover the deployed system exactly (Section~\ref{sec:prereg}), so we checked it in four ways.

\paragraph{Left-out vendors and countries} With each of the 21 vendor and country groups that contain a disputed video removed from calibration in turn, the certified rule falsely confirmed 12 (frame) and 10 (temporal) of the 238 disputed group memberships in the left-out groups (0.050 and 0.042; Table~\ref{tab:validity}). Standard calibration falsely confirmed 49 and 51 (0.21), and group-wise conformal calibration 23 and 17. Because a disputed video counts once for its country and once for its recorded vendor, pooled memberships are not independent, and we give no interval for them. For the six certified groups alone (145 disputed group memberships), the certified rule falsely confirmed 2 (frame; Fig.~\ref{fig:price}b) and 5 (temporal), against 26 and 27 under standard calibration; for one vendor the standard rule reached 12 of 32 with the temporal scorer.

\paragraph{The test set, with clustering} On the 300 test videos (Section~\ref{sec:res-test}), country-clustered bootstrap intervals for disputed false confirmation were 0.067--0.333 for the standard rule and 0--0.105 for the certified rule (frame scorer), and 0.158--0.452 and 0--0.152 for the temporal scorer (both post hoc).

\paragraph{Split calibration (post hoc)} We calibrated each of the five pipelines only on the out-of-fold scores of the 140 development videos it never saw (23--34 disputed videos each) and applied it to the test videos routed to it. Vendor and country groups are too small within a fold, so only the overall, disputed and premature risks were certified. This removes the cross-fitting approximation: each pipeline's threshold carries the guarantee of Theorem~\ref{thm:valid} for its own risks, and the five hold together with probability at least $1 - 5\delta$ by the union bound, while label shift and clustering still apply. On the test set the standard rule then falsely confirmed 8 of 39 disputed videos and the certified rule 2, with overall false confirmation 3/238, no premature confirmation, and timely sensitivity 4/62.

\paragraph{Resampling} Over 1,000 bootstrap calibrations on the development population, the certified rule broke some certified limit in 1.6\,\% (frame) and 0.2\,\% (temporal) of calibrations, below $\delta = 10\,\%$ (Table~\ref{tab:validity}). Standard and PAC marginal calibration broke the disputed limit in essentially every calibration. Because the population is the development set itself, this is an illustration of the two kinds of control, not independent evidence of validity.

\subsection{Protection in expectation or with high probability}\label{sec:res-groupwise}

Group-wise conformal calibration removed most of the disputed excess on development, with higher timely sensitivity than the certified rule (Table~\ref{tab:dev}). It protects each group only in expectation, so a given calibrated system exceeded some limit in 19--32\,\% of resampled calibrations. On the test set (post hoc), it held disputed false confirmation at 2/39 with the frame scorer and confirmed 11 of 62 sufficient videos on time, against 6 under the certified rule. With the temporal scorer it confirmed 6 of 39 disputed videos (0.154), above $\alpha$, where the certified rule confirmed 3. Group awareness removes the systematic failure; certification adds assurance for the particular calibrated system, at a further cost in sensitivity.

\subsection{What protection costs}\label{sec:res-price}

\begin{table}[!tb]
\centering
\caption{Clinical read-out per 100 videos (post hoc on test; development cross-fitted). Conf.: confirmations; False: false confirmations; Disp.: false confirmations of disputed videos; PPV: positive predictive value of a confirmation (count).}\label{tab:readout}
\footnotesize
\setlength{\tabcolsep}{3pt}
\begin{tabular}{lllcccc}
\toprule
Data & Scorer & Rule & Conf. & False & Disp. & PPV \\
\midrule
Test & frame & standard & 18.0 & 8.3 & 2.7 & 0.54 (29/54) \\
     &       & certified & 5.0 & 2.3 & 0.7 & 0.53 (8/15) \\
     & temporal & standard & 18.3 & 8.7 & 4.0 & 0.53 (29/55) \\
     &          & certified & 6.3 & 2.0 & 1.0 & 0.68 (13/19) \\
\midrule
Dev. & frame & standard & 21.6 & 7.0 & 4.0 & 0.68 (102/151) \\
     &       & certified & 5.4 & 1.1 & 1.0 & 0.79 (30/38) \\
     & temporal & standard & 21.3 & 7.1 & 4.0 & 0.66 (99/149) \\
     &          & certified & 8.3 & 1.7 & 0.9 & 0.79 (46/58) \\
\bottomrule
\end{tabular}
\end{table}

\paragraph{Sensitivity} Fig.~\ref{fig:price}a shows the trade-off on the test set (post hoc). Read off the test curve in hindsight, the best sensitivity any threshold on the frame scorer reached at the disputed limit was 0.24, consistent with its modest separability of sufficient from disputed videos (case AUROC 0.69 on both development and test) and with the bound of Proposition~\ref{prop:price}. The certified rule achieved 0.13. The difference reflects the margin needed to hold the limit with high probability from 135 disputed calibration videos, the other certified risks, and sampling noise and label shift between development and test. For the temporal scorer, the certified rule reached the best attainable value on its test curve (0.21; Table~\ref{tab:endpoints}, E6). Under the fitted panel model an ideal recognizer would reach about 0.66 at the same protection, but this ceiling is not identified (Section~\ref{sec:price}).

\paragraph{What a confirmation means} Per 100 test videos, the certified rule gave 5.0 confirmations with the frame scorer, of which 2.3 were false and 0.7 concerned disputed videos, against 18.0, 8.3 and 2.7 under standard calibration (Table~\ref{tab:readout}). The certified rule controls false confirmation among insufficient videos, not the positive predictive value of a confirmation, which also depends on prevalence and sensitivity. With 21\,\% sufficient videos in the test set, about half of the confirmations of either rule with the frame scorer were of insufficient videos, most of them clear (5 of 7 under the certified rule). Certifying the false-discovery rate among confirmations instead was not feasible: on development, no threshold on either scorer could be certified at a false-discovery rate of 0.3 or less.

\subsection{Sensitivity analyses}\label{sec:res-sens}

Under the vote-count reference of Assumption~\ref{ass:panel} (sufficient if $V \ge 2$, disputed if $V = 1$), standard calibration falsely confirmed 12 of 46 disputed test videos (frame) and 17 of 46 (temporal), and the certified rule 4 and 5; on development, both scorers gave 0.205 under the standard rule and 0.048 under the certified rule. Among sufficient videos with an observed onset of CVS, premature confirmation on development was 0.225 (standard) against 0.053 (certified) for the frame scorer and 0.219 against 0.066 for the temporal scorer; on test it was 7/38 against 2/38 and 6/38 against 0/38 (post hoc). At other levels ($\alpha = 0.05$ and $0.20$; Table~\ref{tab:alpha}) the same ordering held. At $\alpha = 0.05$ the certified rule confirmed 7--9\,\% of sufficient videos on time, and $\alpha = 0.01$ cannot be certified with 135 disputed videos (Lemma~\ref{lem:nmin}).

\section{Discussion}\label{sec:discussion}

\paragraph{Why accuracy does not buy safety} The paradox follows from how the reference standard is made, not from a particular model. A video that one of three raters judged sufficient is, in the expert population, one that a substantial minority of experts would pass (Fig.~\ref{fig:method}d). A recognizer that learns what experts see scores such videos higher than videos nobody would pass. Under an average guarantee, the false-confirmation budget is spent on the highest-scoring insufficient videos, and the better the recognizer, the more of that budget goes to disputed videos (Theorems~\ref{thm:paradox}--\ref{thm:monotone}). Better recognition cannot remove this: by Theorem~\ref{thm:paradox}, even the ideal recognizer needs explicit protection. Training on panel disagreement is no shortcut either, because which raters sit on a panel, and so whether a given video ends up disputed, is a sampling outcome that the video does not determine, which makes disagreement a noisy target. Our attempts to learn an ambiguity signal or boundary-aware scores did not separate disputed from sufficient videos better than the frame scorer (\ref{app:negative}); under the fitted model the ideal recognizer would separate them far better (AUROC 0.87), so this limit lies in our recognizers, not in the votes. The remedy for the paradox is to decide differently: protect the disputed group explicitly.

\paragraph{Beyond CVS} Nothing in Theorems~\ref{thm:paradox}--\ref{thm:monotone} is specific to surgery. Whenever a binary decision is calibrated against the majority of a small panel drawn from a larger pool of readers (Assumption~\ref{ass:panel}), or of a fixed panel of readers of different strictness (Remark~\ref{rem:rasch}), as in lung-nodule annotation by four of twelve radiologists \citep{armato2011lidc}, chest-radiograph benchmarks with radiologist majorities \citep{irvin2019chexpert} or graded retinal images \citep{krause2018grader}, the stratum where the panel splits is the one in which an average guarantee fails, and recognizers that recover the readers' judgement with less noise fail there more. Medical image analysis benchmarks that rank methods by accuracy alone will not reveal this. The certification step requires only the votes and enough disputed calibration cases (Lemma~\ref{lem:nmin}).

\paragraph{What certification delivers} Certified confirmation protects three groups for clinical reasons: disputed videos, because a confirmation there would mislead; premature confirmation, because a confirmation must rest on what is visible; and vendor and country groups, because a check deployed across hospitals should not fail in one of them. It needs disputed labels only to calibrate and applies to any scorer as post-processing. On a pre-specified multi-centre test it showed no evidence of exceeding its limits; it kept them for vendor and country groups left out of calibration, and a split calibration without cross-fitting gave the same disputed result on test. These are the empirical checks that its approximate exchangeability requires. Its cost is sensitivity: on test it confirmed 10\,\% of sufficient videos on time with the primary scorer, where the group-wise conformal rule confirmed 18\,\% with weaker assurance.

\paragraph{More than a higher threshold} A reviewer may ask whether certification is just a higher threshold. At use it is: one threshold for every video, which is what allows a group that cannot be observed at the bedside to be protected. What certification adds is how high, with a guarantee. Thresholds raised by convention did not hold the disputed limit (a probability of 0.5 gave 0.148 and 0.267 on development), the scan stops at the most permissive threshold that holds every limit, and Lemma~\ref{lem:nmin} says when no threshold can. The conservativeness is set by the recognizer, not by the rule: no rule can exceed the separability bound of Proposition~\ref{prop:price}, the certified temporal scorer reached the best sensitivity attainable on its test curve at the disputed limit (0.21), and Corollary~\ref{cor:gain} shows that, in the panel model, a recognizer held to the disputed limit gains sensitivity rather than disputed errors as its noise falls. The low sensitivity we report is therefore the honest price of the declared limit with today's recognizers.

\paragraph{What this study shows, and what it does not} It shows a failure mode of calibrated confirmation under panel-derived labels, explains it, and gives a remedy whose guarantee was tested on a held-out multi-centre set. It does not show that a certified check improves surgical decisions or reduces injury: the target is agreement with an independent expert majority, the recognizers are modest, and no surgeon used the system. A clinically useful check will need stronger recognizers, which separate sufficient from disputed videos better, and a prospective evaluation of how surgeons respond to a confirmation and to ``not yet''.

\paragraph{Clinical use} We see the rule as a conservative second check and a documentation aid, not as automation. It does not replace verification: a surgeon who must establish CVS anyway gains little from a rare green light, and frequent ``not yet'' signals risk alarm fatigue. Its value is asymmetric. A confirmation is protected even for disputed videos, so the system rarely endorses the cases on which experts themselves disagree. This matters because of automation bias \citep{goddard2012automation}: an unprotected confirming system would be most persuasive precisely where it is least reliable. A confirmation is nonetheless not proof of CVS. About half of test confirmations were of insufficient videos, so a confirmation is an advisory signal that should prompt, not replace, the surgeon's own check; it does not authorize clipping or division. Criterion guidance was weak (\ref{app:guidance}) and rarely flagged every missing criterion of a disputed video. Finally, $\alpha$ is a clinical choice. A level of 0.10 is lax for a check before clipping, but stricter levels need more disputed calibration cases (Lemma~\ref{lem:nmin}).

\paragraph{Implications for benchmarks and datasets} Benchmarks should report decision-level risks within the disputed stratum alongside mAP, because a higher mAP can come with a less safe decision rule. Datasets should keep the individual votes: a certificate for a stratum needs at least $\lceil\log\delta/\log(1-\alpha)\rceil$ calibration cases in it, and datasets that release only consensus labels, or have few disputed cases, cannot support one. Larger panels would identify the protected-sensitivity ceiling, which three votes cannot, and shrink the share of cases whose reference status is itself uncertain.

\paragraph{Limitations} The theory idealizes the panel as an exchangeable draw and uses a vote-count reference; SAGES allocated raters automatically, without describing the allocation as random, and its reference uses per-criterion majorities. Theorem~\ref{thm:monotone} covers recognizers that are noisy versions of the ideal one, and the empirical association across scorers is consistent with it but uncertain once resampling is accounted for. Quantities such as the share of disputed confirmations that are wrong by the expert population are model-based and not identified. The guarantee assumes independent, identically distributed cases and a scorer fitted on separate data; our calibration is cross-fitted, cases cluster by centre, and the training and test sets differ in label distribution, so the guarantee is approximate and our checks of it are empirical. It holds for the population of vendors and countries represented, not for each future site, and covers the 90-s window before clipping; continuous monitoring over a full dissection would need calibration on matched windows (Proposition~\ref{prop:monitor}). The test set held only 39 disputed videos, so E1b has low power and the vendor-0 estimate (0.139) shows the variability of small groups. The temporal scorer depends indirectly on development labels and was selected on development data, so its results are exploratory. The Endoscapes2023 replication rests on one lenient annotator. Finally, protection is bought with sensitivity, and the protected sensitivity of our scorers is far below what a better recognizer might reach; whether stronger recognizers narrow this gap is for future work.

\section{Conclusion}\label{sec:conclusion}

When labels come from sampled expert panels, a CVS recognizer calibrated for average safety concentrates its false confirmations on the videos that some experts would pass, and for recognizers that recover expert judgement with less noise, the concentration grows with accuracy. We proved this accuracy--safety paradox, characterized what it costs to remove it and which part of that cost the data identify, and found its signature in development data and in a pre-specified multi-centre test, where standard calibration exceeded the disputed limit. Certified confirmation is designed to remove it: under exchangeability it bounds disputed, premature and vendor and country false confirmation simultaneously with high probability, using disagreement labels only to calibrate; its protection held on left-out groups, and the test showed no evidence of violation. In the panel model, once the disputed group is protected, lower recognition noise buys sensitivity instead of hidden risk. Before an automated CVS check, or any decision rule trained on panel labels, is trusted, it should be held to guarantees of this kind within the disputed stratum, not only to an accuracy ranking.

\appendix
\renewcommand{\thesection}{Appendix~\Alph{section}}

\section{Proofs}\label{app:proofs}

\paragraph{Reduction to $Q$} Under Assumption~\ref{ass:panel}, $\P(\phi(X) = 1 \mid V = v) = \E[\E[\phi(X) \mid Q] \mid V = v]$, because $V$ is conditionally independent of $X$ given $Q$. Every video-based rule therefore has the operating characteristics of a randomized rule $\psi(Q)$, and since $Q = q(X)$, every threshold rule on $Q$ is a video-based rule.

\paragraph{Monotone likelihood ratio} For the hypergeometric distribution,
\[
\frac{\P(V = v+1 \mid Q)}{\P(V = v \mid Q)} = \frac{(Q - v)(m - v)}{(v+1)(J - Q - m + v + 1)}
\]
wherever both probabilities are positive. This ratio is non-decreasing in $Q$, so $\P(V = l \mid Q)/\P(V = j \mid Q)$ is non-decreasing in $Q$ for $j < l$ on the support. For $Q > J - m + h$, $\P(V \le h \mid Q) = 0$, and such $Q$ carry no insufficient mass.

\paragraph{Proof of Theorem~\ref{thm:paradox}} \emph{Part (i).} The likelihood ratio of $Q$ for ``sufficient'' against ``insufficient'' is proportional to $A(q)/B(q)$, with $A(q) = \sum_{l > h} a_l(q)$, $B(q) = \sum_{j \le h} b_j(q)$, and $a_l(q) = b_l(q) = \P(V = l \mid q)$. For $q < q'$,
\begin{align*}
&A(q')B(q) - A(q)B(q') \\
&\quad= \sum_{l > h}\sum_{j \le h} \bigl[a_l(q') b_j(q) - a_l(q) b_j(q')\bigr] \ge 0,
\end{align*}
because each $a_l/b_j$ is non-decreasing. So $A/B$ is non-decreasing. By the Neyman--Pearson lemma, a threshold rule on $q$, randomized at the boundary, is a most powerful level-$\alpha$ test. The level constraint fixes the threshold and the randomization, so this rule is $\phi^\star$.
\emph{Part (ii).} $\P(V = h \mid q)/\P(V \le h \mid q) = \bigl(\sum_{j \le h} \P(V = j \mid q)/\P(V = h \mid q)\bigr)^{-1}$ is non-decreasing, because each term with $j < h$ is non-increasing. The same $\phi^\star$ is therefore most powerful at the same level. Where the likelihood ratio is flat, $\phi^\star$ is one of several most powerful tests, which suffices.
\emph{Part (iii).} The monotone likelihood ratio implies that the law of $Q$ given $V = h$ stochastically dominates that given $V \le h$, which dominates that given $V < h$. Since $\phi^\star$ is non-decreasing in $Q$, the inequalities follow.

\paragraph{Proof of Theorem~\ref{thm:monotone}} \emph{Composition.} The kernel $f_\sigma(s \mid q) \propto \exp\{-(s - \eta(q))^2/2\sigma^2\}$ is totally positive of order 2 (TP$_2$) in $(q, s)$, because $\eta$ is increasing. $\P(V = v \mid q)$ is TP$_2$ in $(q, v)$, so by Bayes' rule $\P(Q = q \mid V = v)$ is TP$_2$ in $(v, q)$. By the composition theorem for TP$_2$ kernels \citep{karlin1968total}, $f_\sigma(s \mid v) = \sum_q f_\sigma(s \mid q)\,\P(Q = q \mid V = v)$ is TP$_2$ in $(v, s)$: $S_\sigma$ has a monotone likelihood ratio in $V$.
\emph{Most powerful tests.} By the argument of Theorem~\ref{thm:paradox}, the likelihood ratios of ``sufficient'' and of ``disputed'' against ``insufficient'' are non-decreasing in $s$. The level-$\alpha$ threshold rule on $S_\sigma$ is therefore most powerful for both, based on $S_\sigma$.
\emph{Garbling.} For $\sigma' > \sigma$, $S_{\sigma'}$ has the law of $S_\sigma + (\sigma'^2 - \sigma^2)^{1/2} Z'$ with independent $Z'$. Any rule based on $S_{\sigma'}$ is thus a randomized rule based on $S_\sigma$, and the power of the most powerful test based on $S_\sigma$ is at least that based on $S_{\sigma'}$ \citep{blackwell1953equivalent}, at every level. This gives (ii), (iii), and (i) through the ROC curve.
\emph{Part (iv)} follows from Corollary~\ref{cor:conservation} at fixed $\FR_{\all} = \alpha$.
\emph{Limits.} As $\sigma \to \infty$, $S_\sigma$ becomes independent of $V$ and every group's rate tends to $\alpha$. As $\sigma \to 0$, the threshold rule converges to $\phi^\star$.

\paragraph{Corollary~\ref{cor:conservation}} Insufficient cases are disputed or clear, so $\FR_{\all} = \pi\FR_{\dis} + (1-\pi)\FR_{\clr}$.

\paragraph{Remark~\ref{rem:rasch}} For a fixed panel $\mathcal{A}$ with votes $u_e \in \{0, 1\}$, under local independence,
\[
\P(u \mid \theta) = \frac{\exp\bigl(\theta \sum_{e \in \mathcal{A}} u_e - \sum_{e \in \mathcal{A}} b_e u_e\bigr)}{\prod_{e \in \mathcal{A}}(1 + e^{\theta - b_e})}.
\]
Summing over vote vectors with $\sum_e u_e = v$ gives $\P(V = v \mid \theta) = c_v e^{\theta v}/C(\theta)$, an exponential family in $\theta$, which has a monotone likelihood ratio.

\paragraph{Proof of Corollary~\ref{cor:gain}} By the composition step in the proof of Theorem~\ref{thm:monotone}, $f_\sigma(s \mid v)$ is TP$_2$ in $(v, s)$. The likelihood ratio of ``sufficient'' against ``disputed'' based on $S_\sigma$ is $\sum_{v > h} w_v\, f_\sigma(s \mid v)/f_\sigma(s \mid h)$ with weights $w_v \geq 0$, and each term is non-decreasing in $s$, so the threshold rule $\phi_\sigma$ is most powerful at disputed level $\alpha$ among rules based on $S_\sigma$ (Neyman--Pearson; $S_\sigma$ is continuous, so no randomization is needed). Its sensitivity is therefore $\beta_{S_\sigma}(\alpha)$. Because the law of $S_\sigma$ given $V = v$ is stochastically non-decreasing in $v$, the overall false confirmation $\P(S_\sigma \geq t \mid V \leq h)$, a mixture over $v \leq h$, is at most the disputed one, $\alpha$. For $\sigma' > \sigma$, $S_{\sigma'}$ is a garbling of $S_\sigma$, so $\beta_{S_{\sigma'}}(\alpha) \leq \beta_{S_\sigma}(\alpha)$ \citep{blackwell1953equivalent}.

\paragraph{Proof of Proposition~\ref{prop:price}} A rule with disputed false confirmation at most $\alpha$ is a level-$\alpha$ test of ``disputed'' against ``sufficient''. Its power is at most the power of the most powerful test based on its information. A threshold rule on a scalar score with disputed false-positive rate at most $\alpha$ has sensitivity at most the score's ROC value at $\alpha$, by definition. Under Assumption~\ref{ass:panel}, the reduction to $Q$ above gives $\beta_X \le \beta_Q$.

\paragraph{Identification bounds} Let $p_v$ be the observed vote-count shares and $H_{qv} = \P(V = v \mid Q = q)$. The laws $\mu$ of $Q$ consistent with the data satisfy $\sum_q \mu_q H_{qv} = p_v$ for every $v$. The protected sensitivity of the ideal recognizer is $\max_{0 \le u_q \le \mu_q} \sum_q u_q \P(V \ge 2 \mid q) / \sum_{v \ge 2} p_v$ subject to $\sum_q u_q H_{q1} \le \alpha p_1$. Its upper bound, maximizing jointly over $(\mu, u)$, is a linear program. For its lower bound, linear-programming duality writes the inner maximum as $\min_{\lambda \ge 0} [\lambda \alpha p_1 + \sum_q \mu_q (\P(V \ge 2 \mid q) - \lambda H_{q1})^+] / \sum_{v \ge 2} p_v$, and minimizing over $\mu$ for each $\lambda$ on a grid is again a linear program. The ideal recognizer's disputed false confirmation at marginal level $\alpha$ (constraint $\sum_q u_q (H_{q0} + H_{q1}) \le \alpha (p_0 + p_1)$, objective $\sum_q u_q H_{q1} / p_1$) and the share of disputed videos with $Q/J < 1/2$ (objective $\sum_{q < J/2} \mu_q H_{q1}/p_1$) are bounded in the same way. Bounds for $J \to \infty$ (a fine grid of binomial kernels) were similar: ceiling 0.49--0.87 in training, disputed false confirmation 0.18--0.29, share 0.41--1.00. All bounds treat the observed vote shares as exact, and the lower bounds that minimize over a grid of $\lambda$ are approximate.

\paragraph{Proof of Proposition~\ref{prop:monitor}} $\{\exists t \le T: w_t \ge \tau\} = \{E \ge \tau\}$, and likewise with $t < t^f$ and $E^{\pre}$.

\paragraph{Proof of Lemma~\ref{lem:nmin}} The smallest attainable $p$-value is $(1-\alpha)^{n_g}$, attained with no event among the group's calibration cases. If it exceeds $\delta$ for some certified group, no threshold is certified and $\hat\tau = +\infty$.

\paragraph{Proof of Theorem~\ref{thm:valid}} Write $R_g(\tau) = \P(T_g \ge \tau \mid i \in g)$ for a certified risk and $F$ for the $\mathrm{Bin}(n_g, \alpha)$ distribution function. Given the membership of group $g$, its calibration cases are independent and identically distributed, so $N_g(\tau) \sim \mathrm{Bin}(n_g, R_g(\tau))$. $R_g$ and $N_g$ are non-increasing in $\tau$, and $R_g$ is left-continuous, so the violation set $\mathcal{V}_g = \{\tau : R_g(\tau) > \alpha\}$ is a half-line $(-\infty, \tau_g^\ast)$ or $(-\infty, \tau_g^\ast]$. These half-lines are nested; let $g_0$ have the largest. An error means that some $\tau \in \mathcal{V}_{g_0}$ is certified, which requires $F(N_{g_0}(\tau)) \le \delta$. With $K = \inf_{\tau \in \mathcal{V}_{g_0}} N_{g_0}(\tau) = \#\{i : T_{g_0,i} \ge \tau^\ast_{g_0}\}$, an error implies $F(K) \le \delta$. Now $K \sim \mathrm{Bin}(n_{g_0}, r)$ with $r = R_{g_0}(\tau^\ast_{g_0}) \ge \alpha$ by left-continuity, hence $\P(\text{error}) \le \P(F(K) \le \delta) \le \delta$. The bound covers every certified threshold, whatever the candidate list.

\section{Example selection}\label{app:examples}
The disputed example (Figs.~\ref{fig:overview}b and~\ref{fig:method}a) is the disputed training video with the highest case statistic among those confirmed by the standard rule and not by the certified rule (frame scorer). The sufficient example (Fig.~\ref{fig:overview}a,b) is the sufficient video, confirmed on time by the certified rule, whose delay after the first majority-CVS frame is closest to the median (6\,s), with ties broken by video identifier. The clear example (Fig.~\ref{fig:overview}b) is the clear insufficient video whose case statistic is closest to the median among clear videos. Both rules use the thresholds calibrated on all 700 training videos.

\section{Method variants that did not improve protected sensitivity}\label{app:negative}
All variants were evaluated on SAGES training data by cross-fitting, before the lock. An ambiguity-adaptive threshold, which raised the threshold where a head trained to predict rater disagreement expected it, lowered protected sensitivity relative to a single protected threshold ($-0.039$; 95\,\% CI $-0.068$ to $-0.015$; frozen features); predicted disagreement did not separate sufficient from disputed videos (case AUROC 0.63). Training a video-level ordinal score on the number of raters raised sensitivity under disputed protection from 0.13 to 0.23 relative to the frozen scorer, but mainly by anticipating CVS: its timely sensitivity rose only from 0.11 to 0.13, and under standard calibration it confirmed 18\,\% of sufficient videos prematurely, against 8\,\% for the frozen scorer. This motivated certifying premature confirmation. Persistence-adaptive certification over windows of 3--20\,s changed timely sensitivity by $+0.024$ ($-0.022$ to $+0.071$). Cumulative-sum (CUSUM) quickest detection \citep{page1954continuous} in place of the running maximum lowered timely sensitivity ($-0.049$ with the temporal scorer). The boundary-aware temporal model, the originally pre-specified scorer, added ordinal and boundary-ranking losses to the temporal head and gained $+0.010$ ($-0.036$ to $+0.056$) over the frame scorer, below its decision rule. None improved the separation of sufficient from disputed videos (development case AUROC 0.63--0.71), so none could raise the bound of Proposition~\ref{prop:price}.

\section{Criterion guidance}\label{app:guidance}
When the rule does not confirm, it can tell the surgeon which criterion is not yet demonstrated. For criterion $c$, let $M_{i,c}$ be the monitored statistic of its own logit stream (the running maximum of its 3-s minimum). The rule alerts ``criterion $c$ not yet demonstrated'' when $M_{i,c} < \theta_c$, where $\theta_c$ is the largest value, scanned upwards, at which false alerts among calibration videos where the criterion is achieved are certified at most $\alpha$, with confidence $1-\delta$, by the same binomial test. On test, the frame scorer's false-alert rates were 0.062, 0.083 and 0.087 for C1--C3, with recall of missing criteria 0.43--0.57. The temporal scorer's were 0.062, 0.107 and 0.120 (C2 and C3 above $\alpha$, not significantly; $p \ge 0.31$), with recall 0.36--0.44. Every missing criterion was flagged in 15\,\% of disputed videos.

\section{Additional tables}\label{app:tables}
\setcounter{table}{0}
\renewcommand{\theHtable}{app.\thesection.\arabic{table}}

\begin{table}[!htb]
\centering
\caption{Ablation of the certified family (SAGES development, cross-fitted; after unblinding): timely sensitivity and disputed false confirmation (FR) when only some risks are certified.}\label{tab:ablation}
\footnotesize
\setlength{\tabcolsep}{3pt}
\begin{tabular}{lcccc}
\toprule
 & \multicolumn{2}{c}{Frame} & \multicolumn{2}{c}{Temporal} \\
\cmidrule(lr){2-3}\cmidrule(lr){4-5}
Certified risks & Timely & FR & Timely & FR \\
\midrule
overall only & 0.332 & 0.178 & 0.312 & 0.170 \\
overall + disputed & 0.107 & 0.052 & 0.200 & 0.059 \\
overall + premature & 0.205 & 0.111 & 0.195 & 0.059 \\
overall + groups & 0.176 & 0.111 & 0.176 & 0.044 \\
all (certified rule) & 0.107 & 0.052 & 0.176 & 0.044 \\
\bottomrule
\end{tabular}
\end{table}

\begin{table}[!htb]
\centering
\caption{Other levels $\alpha$ (SAGES development, cross-fitted, $\delta = 0.1$; after unblinding): disputed false confirmation (FR) and timely sensitivity.}\label{tab:alpha}
\footnotesize
\setlength{\tabcolsep}{3pt}
\begin{tabular}{llcccc}
\toprule
 & & \multicolumn{2}{c}{Frame} & \multicolumn{2}{c}{Temporal} \\
\cmidrule(lr){3-4}\cmidrule(lr){5-6}
$\alpha$ & Rule & FR & Timely & FR & Timely \\
\midrule
0.05 & standard & 0.126 & 0.234 & 0.089 & 0.254 \\
     & group-wise & 0.037 & 0.112 & 0.044 & 0.137 \\
     & certified & 0.022 & 0.088 & 0.015 & 0.073 \\
\midrule
0.20 & standard & 0.393 & 0.351 & 0.393 & 0.341 \\
     & group-wise & 0.200 & 0.341 & 0.185 & 0.317 \\
     & certified & 0.141 & 0.283 & 0.156 & 0.307 \\
\bottomrule
\end{tabular}
\end{table}

\begin{table}[!htb]
\centering
\caption{Standard calibration on Endoscapes2023 development cases ($\alpha = 0.1$, cross-fitted; hypothesis-generating), seven recognizers ordered by frame mAP: false confirmations (FC) of disputed (of 20) and clear (of 89) cases, and case AUROC of sufficient against all insufficient (the x-axis of Fig.~\ref{fig:paradox}c), disputed and clear cases. SA: structure-aware; DA: decision-aligned; FT: fine-tuned (last 4 blocks); DI: detector-initialized.}\label{tab:replication}
\footnotesize
\setlength{\tabcolsep}{2.5pt}
\begin{tabular}{lcccccc}
\toprule
 & & \multicolumn{2}{c}{FC} & \multicolumn{3}{c}{AUROC vs} \\
\cmidrule(lr){3-4}\cmidrule(lr){5-7}
Recognizer & mAP & disp. & clear & all & disp. & clear \\
\midrule
ResNet-18, frozen & 0.284 & 3 & 8 & 0.54 & 0.49 & 0.55 \\
DINOv3-B, frozen & 0.384 & 3 & 7 & 0.68 & 0.59 & 0.70 \\
DINOv3-B, SA + DA & 0.392 & 4 & 7 & 0.71 & 0.61 & 0.73 \\
DINOv2-B, frozen & 0.401 & 6 & 6 & 0.66 & 0.52 & 0.69 \\
DINOv3-B, SA & 0.420 & 4 & 5 & 0.71 & 0.59 & 0.73 \\
DINOv3-B, FT & 0.429 & 6 & 4 & 0.72 & 0.54 & 0.77 \\
ResNet-50, DI & 0.439 & 6 & 6 & 0.71 & 0.58 & 0.74 \\
\bottomrule
\end{tabular}
\end{table}

\FloatBarrier

\section*{CRediT authorship contribution statement}
\textbf{Bowen Xiang:} Conceptualization, Methodology, Software, Formal analysis, Investigation, Data curation, Visualization, Writing -- original draft, Writing -- review \& editing, Funding acquisition. \textbf{Xiao Yu:} Validation, Writing -- review \& editing. \textbf{Mingze Xu:} Validation, Writing -- review \& editing.

\section*{Declaration of competing interests}
The authors declare that they have no known competing financial interests or personal relationships that could have appeared to influence the work reported in this paper.

\section*{Funding}
This work was supported by the Vanderbilt University Innovation Catalyst Fund. The funder had no role in study design, analysis, interpretation or the decision to publish.

\section*{Acknowledgements}
We thank the organizers, annotators and contributing centres of the SAGES CVS Challenge, Endoscapes2023 and Cholec80-CVS for making their data public.

\section*{Ethics statement}
Only public, de-identified surgical video datasets were used, under their licences. No new data were collected, and no patient-identifying information is included.

\section*{Data and code availability}
The video datasets are public and available from their providers: SAGES CVS Challenge 2024 (CC BY-NC 4.0), Endoscapes2023 (CC BY-NC-SA 4.0) and Cholec80-CVS \citep{alapatt2026sages,murali2023endoscapes,rios2023cholec80cvs}. The code and the derived data (the pre-specification, the lock file with all hashes, the test-access log, per-video scores and thresholds, and the scripts behind every number and figure) will be made publicly available upon publication of the peer-reviewed version. The frames in Figs.~\ref{fig:overview} and \ref{fig:method} are reproduced from SAGES training videos under CC BY-NC 4.0.

\section*{Declaration of generative AI and AI-assisted technologies in the manuscript preparation process}
During the preparation of this work the authors used Claude Code (Anthropic) to assist with code implementation and with the writing and language editing of the manuscript. After using this tool, the authors reviewed and edited the content as needed and take full responsibility for the content of the published article.

\bibliographystyle{elsarticle-harv}
\bibliography{refs}

\begin{thebibliography}{67}
\expandafter\ifx\csname natexlab\endcsname\relax\def\natexlab#1{#1}\fi
\providecommand{\url}[1]{\texttt{#1}}
\providecommand{\href}[2]{#2}
\providecommand{\path}[1]{#1}
\providecommand{\DOIprefix}{doi:}
\providecommand{\ArXivprefix}{arXiv:}
\providecommand{\URLprefix}{URL: }
\providecommand{\Pubmedprefix}{pmid:}
\providecommand{\doi}[1]{\href{http://dx.doi.org/#1}{\path{#1}}}
\providecommand{\Pubmed}[1]{\href{pmid:#1}{\path{#1}}}
\providecommand{\bibinfo}[2]{#2}
\ifx\xfnm\relax \def\xfnm[#1]{\unskip,\space#1}\fi
\bibitem[{Alapatt et~al.(2026)Alapatt, Eckhoff, Lyu, Ban, Mazellier, Choksi,
  Yang, Chiang, Zorzetti, Cannas, Neimark, Bar, Yamlahi, Hennighausen, Wang,
  Li, Liang, Wang, Koju, Bhattarai, Jaspers, Mao, Wijekoon, Ma, Xu, Weng,
  Okran, Rashwan, Shen, Yang, Zhang, Wang, {{2024 CVS Challenge Consortium}},
  Li, Filicori, Li, Mascagni, Hashimoto, Rosman, Meireles and
  Padoy}]{alapatt2026sages}
\bibinfo{author}{Alapatt, D.}, \bibinfo{author}{Eckhoff, J.},
  \bibinfo{author}{Lyu, Z.}, \bibinfo{author}{Ban, Y.},
  \bibinfo{author}{Mazellier, J.P.}, \bibinfo{author}{Choksi, S.},
  \bibinfo{author}{Yang, K.}, \bibinfo{author}{Chiang, P.H.},
  \bibinfo{author}{Zorzetti, N.}, \bibinfo{author}{Cannas, S.},
  \bibinfo{author}{Neimark, D.}, \bibinfo{author}{Bar, O.},
  \bibinfo{author}{Yamlahi, A.}, \bibinfo{author}{Hennighausen, J.},
  \bibinfo{author}{Wang, X.}, \bibinfo{author}{Li, R.}, \bibinfo{author}{Liang,
  L.}, \bibinfo{author}{Wang, Y.}, \bibinfo{author}{Koju, S.},
  \bibinfo{author}{Bhattarai, B.}, \bibinfo{author}{Jaspers, T.},
  \bibinfo{author}{Mao, Z.}, \bibinfo{author}{Wijekoon, A.},
  \bibinfo{author}{Ma, J.}, \bibinfo{author}{Xu, Y.}, \bibinfo{author}{Weng,
  Z.}, \bibinfo{author}{Okran, A.M.}, \bibinfo{author}{Rashwan, H.A.},
  \bibinfo{author}{Shen, B.}, \bibinfo{author}{Yang, K.},
  \bibinfo{author}{Zhang, Y.}, \bibinfo{author}{Wang, H.},
  \bibinfo{author}{{{2024 CVS Challenge Consortium}}}, \bibinfo{author}{Li,
  Q.}, \bibinfo{author}{Filicori, F.}, \bibinfo{author}{Li, X.},
  \bibinfo{author}{Mascagni, P.}, \bibinfo{author}{Hashimoto, D.A.},
  \bibinfo{author}{Rosman, G.}, \bibinfo{author}{Meireles, O.},
  \bibinfo{author}{Padoy, N.}, \bibinfo{year}{2026}.
\newblock \bibinfo{title}{The {SAGES} {Critical View of Safety} {Challenge}: a
  global benchmark for {AI}-assisted surgical quality assessment}.
\newblock \bibinfo{journal}{Med. Image Anal.} \bibinfo{note}{{In press}.
  {arXiv} preprint arXiv:2509.17100,
  \url{https://doi.org/10.48550/arXiv.2509.17100}}.
\bibitem[{Angelopoulos and Bates(2023)}]{angelopoulos2023gentle}
\bibinfo{author}{Angelopoulos, A.N.}, \bibinfo{author}{Bates, S.},
  \bibinfo{year}{2023}.
\newblock \bibinfo{title}{Conformal prediction: a gentle introduction}.
\newblock \bibinfo{journal}{Found. Trends Mach. Learn.} \bibinfo{volume}{16},
  \bibinfo{pages}{494--591}.
\newblock \DOIprefix\doi{10.1561/2200000101}.
\bibitem[{Angelopoulos et~al.(2025)Angelopoulos, Bates, Cand{\`e}s, Jordan and
  Lei}]{angelopoulos2025learn}
\bibinfo{author}{Angelopoulos, A.N.}, \bibinfo{author}{Bates, S.},
  \bibinfo{author}{Cand{\`e}s, E.J.}, \bibinfo{author}{Jordan, M.I.},
  \bibinfo{author}{Lei, L.}, \bibinfo{year}{2025}.
\newblock \bibinfo{title}{Learn then test: calibrating predictive algorithms to
  achieve risk control}.
\newblock \bibinfo{journal}{Ann. Appl. Stat.} \bibinfo{volume}{19},
  \bibinfo{pages}{1641--1662}.
\newblock \DOIprefix\doi{10.1214/24-AOAS1998}.
\bibitem[{Angelopoulos et~al.(2024a)Angelopoulos, Bates, Fisch, Lei and
  Schuster}]{angelopoulos2024conformal}
\bibinfo{author}{Angelopoulos, A.N.}, \bibinfo{author}{Bates, S.},
  \bibinfo{author}{Fisch, A.}, \bibinfo{author}{Lei, L.},
  \bibinfo{author}{Schuster, T.}, \bibinfo{year}{2024}a.
\newblock \bibinfo{title}{Conformal risk control}, in: \bibinfo{booktitle}{The
  Twelfth International Conference on Learning Representations ({ICLR} 2024)}.
\newblock \URLprefix \url{https://openreview.net/forum?id=33XGfHLtZg}.
\bibitem[{Angelopoulos et~al.(2024b)Angelopoulos, Pomerantz, Do, Bates, Bridge,
  Elton, Lev, Gonz{\'a}lez, Jordan and Malik}]{angelopoulos2024triage}
\bibinfo{author}{Angelopoulos, A.N.}, \bibinfo{author}{Pomerantz, S.R.},
  \bibinfo{author}{Do, S.}, \bibinfo{author}{Bates, S.},
  \bibinfo{author}{Bridge, C.P.}, \bibinfo{author}{Elton, D.C.},
  \bibinfo{author}{Lev, M.H.}, \bibinfo{author}{Gonz{\'a}lez, R.G.},
  \bibinfo{author}{Jordan, M.I.}, \bibinfo{author}{Malik, J.},
  \bibinfo{year}{2024}b.
\newblock \bibinfo{title}{Conformal triage for medical imaging {AI}
  deployment}.
\newblock \bibinfo{howpublished}{medRxiv preprint}.
\newblock \DOIprefix\doi{10.1101/2024.02.09.24302543}.
\bibitem[{{Armato III} et~al.(2011){Armato III}, McLennan, Bidaut, McNitt-Gray,
  Meyer, Reeves et~al.}]{armato2011lidc}
\bibinfo{author}{{Armato III}, S.G.}, \bibinfo{author}{McLennan, G.},
  \bibinfo{author}{Bidaut, L.}, \bibinfo{author}{McNitt-Gray, M.F.},
  \bibinfo{author}{Meyer, C.R.}, \bibinfo{author}{Reeves, A.P.}, et~al.,
  \bibinfo{year}{2011}.
\newblock \bibinfo{title}{The {Lung Image Database Consortium} ({LIDC}) and
  {Image Database Resource Initiative} ({IDRI}): a completed reference database
  of lung nodules on {CT} scans}.
\newblock \bibinfo{journal}{Med. Phys.} \bibinfo{volume}{38},
  \bibinfo{pages}{915--931}.
\newblock \DOIprefix\doi{10.1118/1.3528204}.
\bibitem[{Barber et~al.(2021)Barber, Cand{\`e}s, Ramdas and
  Tibshirani}]{barber2021predictive}
\bibinfo{author}{Barber, R.F.}, \bibinfo{author}{Cand{\`e}s, E.J.},
  \bibinfo{author}{Ramdas, A.}, \bibinfo{author}{Tibshirani, R.J.},
  \bibinfo{year}{2021}.
\newblock \bibinfo{title}{Predictive inference with the jackknife+}.
\newblock \bibinfo{journal}{Ann. Stat.} \bibinfo{volume}{49},
  \bibinfo{pages}{486--507}.
\newblock \DOIprefix\doi{10.1214/20-AOS1965}.
\bibitem[{Bates et~al.(2021)Bates, Angelopoulos, Lei, Malik and
  Jordan}]{bates2021distribution}
\bibinfo{author}{Bates, S.}, \bibinfo{author}{Angelopoulos, A.N.},
  \bibinfo{author}{Lei, L.}, \bibinfo{author}{Malik, J.},
  \bibinfo{author}{Jordan, M.I.}, \bibinfo{year}{2021}.
\newblock \bibinfo{title}{Distribution-free, risk-controlling prediction sets}.
\newblock \bibinfo{journal}{J. ACM} \bibinfo{volume}{68},
  \bibinfo{pages}{1--34}.
\newblock \DOIprefix\doi{10.1145/3478535}.
\bibitem[{Bian and Barber(2023)}]{bian2023training}
\bibinfo{author}{Bian, M.}, \bibinfo{author}{Barber, R.F.},
  \bibinfo{year}{2023}.
\newblock \bibinfo{title}{Training-conditional coverage for distribution-free
  predictive inference}.
\newblock \bibinfo{journal}{Electron. J. Stat.} \bibinfo{volume}{17},
  \bibinfo{pages}{2044--2066}.
\newblock \DOIprefix\doi{10.1214/23-EJS2145}.
\bibitem[{Blackwell(1953)}]{blackwell1953equivalent}
\bibinfo{author}{Blackwell, D.}, \bibinfo{year}{1953}.
\newblock \bibinfo{title}{Equivalent comparisons of experiments}.
\newblock \bibinfo{journal}{Ann. Math. Stat.} \bibinfo{volume}{24},
  \bibinfo{pages}{265--272}.
\newblock \DOIprefix\doi{10.1214/aoms/1177729032}.
\bibitem[{Brunt et~al.(2020)Brunt, Deziel, Telem, Strasberg, Aggarwal, Asbun,
  Bonjer, McDonald, Alseidi, Ujiki, Riall, Hammill, Moulton, Pucher, Parks,
  Ansari, Connor, Dirks, Anderson, Altieri, Tsamalaidze, Stefanidis and
  {{Prevention of Bile Duct Injury Consensus Work Group}}}]{brunt2020safe}
\bibinfo{author}{Brunt, L.M.}, \bibinfo{author}{Deziel, D.J.},
  \bibinfo{author}{Telem, D.A.}, \bibinfo{author}{Strasberg, S.M.},
  \bibinfo{author}{Aggarwal, R.}, \bibinfo{author}{Asbun, H.},
  \bibinfo{author}{Bonjer, J.}, \bibinfo{author}{McDonald, M.},
  \bibinfo{author}{Alseidi, A.}, \bibinfo{author}{Ujiki, M.},
  \bibinfo{author}{Riall, T.S.}, \bibinfo{author}{Hammill, C.},
  \bibinfo{author}{Moulton, C.A.}, \bibinfo{author}{Pucher, P.H.},
  \bibinfo{author}{Parks, R.W.}, \bibinfo{author}{Ansari, M.T.},
  \bibinfo{author}{Connor, S.}, \bibinfo{author}{Dirks, R.C.},
  \bibinfo{author}{Anderson, B.}, \bibinfo{author}{Altieri, M.S.},
  \bibinfo{author}{Tsamalaidze, L.}, \bibinfo{author}{Stefanidis, D.},
  \bibinfo{author}{{{Prevention of Bile Duct Injury Consensus Work Group}}},
  \bibinfo{year}{2020}.
\newblock \bibinfo{title}{Safe cholecystectomy multi-society practice guideline
  and state of the art consensus conference on prevention of bile duct injury
  during cholecystectomy}.
\newblock \bibinfo{journal}{Ann. Surg.} \bibinfo{volume}{272},
  \bibinfo{pages}{3--23}.
\newblock \DOIprefix\doi{10.1097/SLA.0000000000003791}.
\bibitem[{Cho et~al.(2014)Cho, van Merri{\"e}nboer, Gulcehre, Bahdanau,
  Bougares, Schwenk and Bengio}]{cho2014learning}
\bibinfo{author}{Cho, K.}, \bibinfo{author}{van Merri{\"e}nboer, B.},
  \bibinfo{author}{Gulcehre, C.}, \bibinfo{author}{Bahdanau, D.},
  \bibinfo{author}{Bougares, F.}, \bibinfo{author}{Schwenk, H.},
  \bibinfo{author}{Bengio, Y.}, \bibinfo{year}{2014}.
\newblock \bibinfo{title}{Learning phrase representations using {RNN}
  encoder--decoder for statistical machine translation}, in:
  \bibinfo{booktitle}{Proceedings of the 2014 Conference on Empirical Methods
  in Natural Language Processing ({EMNLP})}, \bibinfo{publisher}{Association
  for Computational Linguistics}, \bibinfo{address}{Doha, Qatar}. pp.
  \bibinfo{pages}{1724--1734}.
\newblock \DOIprefix\doi{10.3115/v1/D14-1179}.
\bibitem[{Clopper and Pearson(1934)}]{clopper1934use}
\bibinfo{author}{Clopper, C.J.}, \bibinfo{author}{Pearson, E.S.},
  \bibinfo{year}{1934}.
\newblock \bibinfo{title}{The use of confidence or fiducial limits illustrated
  in the case of the binomial}.
\newblock \bibinfo{journal}{Biometrika} \bibinfo{volume}{26},
  \bibinfo{pages}{404--413}.
\newblock \DOIprefix\doi{10.1093/biomet/26.4.404}.
\bibitem[{Dawid and Skene(1979)}]{dawid1979maximum}
\bibinfo{author}{Dawid, A.P.}, \bibinfo{author}{Skene, A.M.},
  \bibinfo{year}{1979}.
\newblock \bibinfo{title}{Maximum likelihood estimation of observer error-rates
  using the {EM} algorithm}.
\newblock \bibinfo{journal}{J. R. Stat. Soc. Ser. C Appl. Stat.}
  \bibinfo{volume}{28}, \bibinfo{pages}{20--28}.
\newblock \DOIprefix\doi{10.2307/2346806}.
\bibitem[{Dvijotham et~al.(2023)Dvijotham, Winkens, Barsbey, Ghaisas,
  Stanforth, Pawlowski, Strachan, Ahmed, Azizi, Bachrach, Culp, Daswani,
  Freyberg, Kelly, Kiraly, Kohlberger, McKinney, Mustafa, Natarajan, Geras,
  Witowski, Qin, Creswell, Shetty, Sieniek, Spitz, Corrado, Kohli, Cemgil and
  Karthikesalingam}]{dvijotham2023enhancing}
\bibinfo{author}{Dvijotham, K.}, \bibinfo{author}{Winkens, J.},
  \bibinfo{author}{Barsbey, M.}, \bibinfo{author}{Ghaisas, S.},
  \bibinfo{author}{Stanforth, R.}, \bibinfo{author}{Pawlowski, N.},
  \bibinfo{author}{Strachan, P.}, \bibinfo{author}{Ahmed, Z.},
  \bibinfo{author}{Azizi, S.}, \bibinfo{author}{Bachrach, Y.},
  \bibinfo{author}{Culp, L.}, \bibinfo{author}{Daswani, M.},
  \bibinfo{author}{Freyberg, J.}, \bibinfo{author}{Kelly, C.},
  \bibinfo{author}{Kiraly, A.}, \bibinfo{author}{Kohlberger, T.},
  \bibinfo{author}{McKinney, S.}, \bibinfo{author}{Mustafa, B.},
  \bibinfo{author}{Natarajan, V.}, \bibinfo{author}{Geras, K.},
  \bibinfo{author}{Witowski, J.}, \bibinfo{author}{Qin, Z.Z.},
  \bibinfo{author}{Creswell, J.}, \bibinfo{author}{Shetty, S.},
  \bibinfo{author}{Sieniek, M.}, \bibinfo{author}{Spitz, T.},
  \bibinfo{author}{Corrado, G.}, \bibinfo{author}{Kohli, P.},
  \bibinfo{author}{Cemgil, T.}, \bibinfo{author}{Karthikesalingam, A.},
  \bibinfo{year}{2023}.
\newblock \bibinfo{title}{Enhancing the reliability and accuracy of
  {AI}-enabled diagnosis via complementarity-driven deferral to clinicians}.
\newblock \bibinfo{journal}{Nat. Med.} \bibinfo{volume}{29},
  \bibinfo{pages}{1814--1820}.
\newblock \DOIprefix\doi{10.1038/s41591-023-02437-x}.
\bibitem[{Geifman and El-Yaniv(2017)}]{geifman2017selective}
\bibinfo{author}{Geifman, Y.}, \bibinfo{author}{El-Yaniv, R.},
  \bibinfo{year}{2017}.
\newblock \bibinfo{title}{Selective classification for deep neural networks},
  in: \bibinfo{booktitle}{Advances in Neural Information Processing Systems},
  \bibinfo{publisher}{Curran Associates}. pp. \bibinfo{pages}{4878--4887}.
\newblock \URLprefix
  \url{https://proceedings.neurips.cc/paper_files/paper/2017/hash/4a8423d5e91fda00bb7e46540e2b0cf1-Abstract.html}.
\bibitem[{Gibbs et~al.(2025)Gibbs, Cherian and Cand{\`e}s}]{gibbs2025conformal}
\bibinfo{author}{Gibbs, I.}, \bibinfo{author}{Cherian, J.J.},
  \bibinfo{author}{Cand{\`e}s, E.J.}, \bibinfo{year}{2025}.
\newblock \bibinfo{title}{Conformal prediction with conditional guarantees}.
\newblock \bibinfo{journal}{J. R. Stat. Soc. Ser. B Stat. Methodol.}
  \bibinfo{volume}{87}, \bibinfo{pages}{1100--1126}.
\newblock \DOIprefix\doi{10.1093/jrsssb/qkaf008}.
\bibitem[{Goddard et~al.(2012)Goddard, Roudsari and
  Wyatt}]{goddard2012automation}
\bibinfo{author}{Goddard, K.}, \bibinfo{author}{Roudsari, A.},
  \bibinfo{author}{Wyatt, J.C.}, \bibinfo{year}{2012}.
\newblock \bibinfo{title}{Automation bias: a systematic review of frequency,
  effect mediators, and mitigators}.
\newblock \bibinfo{journal}{J. Am. Med. Inform. Assoc.} \bibinfo{volume}{19},
  \bibinfo{pages}{121--127}.
\newblock \DOIprefix\doi{10.1136/amiajnl-2011-000089}.
\bibitem[{Guo et~al.(2017)Guo, Pleiss, Sun and Weinberger}]{guo2017calibration}
\bibinfo{author}{Guo, C.}, \bibinfo{author}{Pleiss, G.}, \bibinfo{author}{Sun,
  Y.}, \bibinfo{author}{Weinberger, K.Q.}, \bibinfo{year}{2017}.
\newblock \bibinfo{title}{On calibration of modern neural networks}, in:
  \bibinfo{booktitle}{Proceedings of the 34th International Conference on
  Machine Learning ({ICML})}, \bibinfo{publisher}{PMLR}. pp.
  \bibinfo{pages}{1321--1330}.
\newblock \URLprefix \url{https://proceedings.mlr.press/v70/guo17a.html}.
\bibitem[{Holm(1979)}]{holm1979simple}
\bibinfo{author}{Holm, S.}, \bibinfo{year}{1979}.
\newblock \bibinfo{title}{A simple sequentially rejective multiple test
  procedure}.
\newblock \bibinfo{journal}{Scand. J. Stat.} \bibinfo{volume}{6},
  \bibinfo{pages}{65--70}.
\newblock \URLprefix \url{https://www.jstor.org/stable/4615733}.
\bibitem[{Irvin et~al.(2019)Irvin, Rajpurkar, Ko, Yu, Ciurea-Ilcus, Chute
  et~al.}]{irvin2019chexpert}
\bibinfo{author}{Irvin, J.}, \bibinfo{author}{Rajpurkar, P.},
  \bibinfo{author}{Ko, M.}, \bibinfo{author}{Yu, Y.},
  \bibinfo{author}{Ciurea-Ilcus, S.}, \bibinfo{author}{Chute, C.}, et~al.,
  \bibinfo{year}{2019}.
\newblock \bibinfo{title}{{CheXpert}: a large chest radiograph dataset with
  uncertainty labels and expert comparison}, in:
  \bibinfo{booktitle}{Proceedings of the AAAI Conference on Artificial
  Intelligence}, pp. \bibinfo{pages}{590--597}.
\newblock \DOIprefix\doi{10.1609/aaai.v33i01.3301590}.
\bibitem[{Jensen et~al.(2019)Jensen, J{\o}rgensen, Jalaboi, Hansen and
  Olsen}]{jensen2019improving}
\bibinfo{author}{Jensen, M.H.}, \bibinfo{author}{J{\o}rgensen, D.R.},
  \bibinfo{author}{Jalaboi, R.}, \bibinfo{author}{Hansen, M.E.},
  \bibinfo{author}{Olsen, M.A.}, \bibinfo{year}{2019}.
\newblock \bibinfo{title}{Improving uncertainty estimation in convolutional
  neural networks using inter-rater agreement}, in: \bibinfo{booktitle}{Medical
  Image Computing and Computer Assisted Intervention -- {MICCAI} 2019},
  \bibinfo{publisher}{Springer}, \bibinfo{address}{Cham}. pp.
  \bibinfo{pages}{540--548}.
\newblock \DOIprefix\doi{10.1007/978-3-030-32251-9_59}.
\bibitem[{Jones et~al.(2021)Jones, Sagawa, Koh, Kumar and
  Liang}]{jones2021selective}
\bibinfo{author}{Jones, E.}, \bibinfo{author}{Sagawa, S.},
  \bibinfo{author}{Koh, P.W.}, \bibinfo{author}{Kumar, A.},
  \bibinfo{author}{Liang, P.}, \bibinfo{year}{2021}.
\newblock \bibinfo{title}{Selective classification can magnify disparities
  across groups}, in: \bibinfo{booktitle}{Ninth International Conference on
  Learning Representations ({ICLR} 2021)}.
\newblock \URLprefix \url{https://openreview.net/forum?id=N0M_4BkQ05i}.
\bibitem[{Jung et~al.(2023)Jung, Noarov, Ramalingam and Roth}]{jung2023batch}
\bibinfo{author}{Jung, C.}, \bibinfo{author}{Noarov, G.},
  \bibinfo{author}{Ramalingam, R.}, \bibinfo{author}{Roth, A.},
  \bibinfo{year}{2023}.
\newblock \bibinfo{title}{Batch multivalid conformal prediction}, in:
  \bibinfo{booktitle}{The Eleventh International Conference on Learning
  Representations ({ICLR} 2023)}.
\newblock \URLprefix \url{https://openreview.net/forum?id=Dk7QQp8jHEo}.
\bibitem[{Karimi et~al.(2020)Karimi, Dou, Warfield and
  Gholipour}]{karimi2020deep}
\bibinfo{author}{Karimi, D.}, \bibinfo{author}{Dou, H.},
  \bibinfo{author}{Warfield, S.K.}, \bibinfo{author}{Gholipour, A.},
  \bibinfo{year}{2020}.
\newblock \bibinfo{title}{Deep learning with noisy labels: exploring techniques
  and remedies in medical image analysis}.
\newblock \bibinfo{journal}{Med. Image Anal.} \bibinfo{volume}{65},
  \bibinfo{pages}{101759}.
\newblock \DOIprefix\doi{10.1016/j.media.2020.101759}.
\bibitem[{Karlin(1968)}]{karlin1968total}
\bibinfo{author}{Karlin, S.}, \bibinfo{year}{1968}.
\newblock \bibinfo{title}{Total positivity}. volume~\bibinfo{volume}{1}.
\newblock \bibinfo{publisher}{Stanford University Press},
  \bibinfo{address}{Stanford, CA}.
\bibitem[{Karlin and Rubin(1956)}]{karlin1956theory}
\bibinfo{author}{Karlin, S.}, \bibinfo{author}{Rubin, H.},
  \bibinfo{year}{1956}.
\newblock \bibinfo{title}{The theory of decision procedures for distributions
  with monotone likelihood ratio}.
\newblock \bibinfo{journal}{Ann. Math. Stat.} \bibinfo{volume}{27},
  \bibinfo{pages}{272--299}.
\newblock \DOIprefix\doi{10.1214/aoms/1177728259}.
\bibitem[{Kawamura et~al.(2023)Kawamura, Endo, Fujinaga, Orimoto, Amano,
  Kawasaki, Kawano, Masuda, Hirashita, Kimura, Ejima, Matsunobu, Shinozuka,
  Tokuyasu and Inomata}]{kawamura2023development}
\bibinfo{author}{Kawamura, M.}, \bibinfo{author}{Endo, Y.},
  \bibinfo{author}{Fujinaga, A.}, \bibinfo{author}{Orimoto, H.},
  \bibinfo{author}{Amano, S.}, \bibinfo{author}{Kawasaki, T.},
  \bibinfo{author}{Kawano, Y.}, \bibinfo{author}{Masuda, T.},
  \bibinfo{author}{Hirashita, T.}, \bibinfo{author}{Kimura, M.},
  \bibinfo{author}{Ejima, A.}, \bibinfo{author}{Matsunobu, Y.},
  \bibinfo{author}{Shinozuka, K.}, \bibinfo{author}{Tokuyasu, T.},
  \bibinfo{author}{Inomata, M.}, \bibinfo{year}{2023}.
\newblock \bibinfo{title}{Development of an artificial intelligence system for
  real-time intraoperative assessment of the critical view of safety in
  laparoscopic cholecystectomy}.
\newblock \bibinfo{journal}{Surg. Endosc.} \bibinfo{volume}{37},
  \bibinfo{pages}{8755--8763}.
\newblock \DOIprefix\doi{10.1007/s00464-023-10328-y}.
\bibitem[{Kompa et~al.(2021)Kompa, Snoek and Beam}]{kompa2021second}
\bibinfo{author}{Kompa, B.}, \bibinfo{author}{Snoek, J.},
  \bibinfo{author}{Beam, A.L.}, \bibinfo{year}{2021}.
\newblock \bibinfo{title}{Second opinion needed: communicating uncertainty in
  medical machine learning}.
\newblock \bibinfo{journal}{npj Digit. Med.} \bibinfo{volume}{4},
  \bibinfo{pages}{4}.
\newblock \DOIprefix\doi{10.1038/s41746-020-00367-3}.
\bibitem[{Krause et~al.(2018)Krause, Gulshan, Rahimy, Karth, Widner, Corrado,
  Peng and Webster}]{krause2018grader}
\bibinfo{author}{Krause, J.}, \bibinfo{author}{Gulshan, V.},
  \bibinfo{author}{Rahimy, E.}, \bibinfo{author}{Karth, P.},
  \bibinfo{author}{Widner, K.}, \bibinfo{author}{Corrado, G.S.},
  \bibinfo{author}{Peng, L.}, \bibinfo{author}{Webster, D.R.},
  \bibinfo{year}{2018}.
\newblock \bibinfo{title}{Grader variability and the importance of reference
  standards for evaluating machine learning models for diabetic retinopathy}.
\newblock \bibinfo{journal}{Ophthalmology} \bibinfo{volume}{125},
  \bibinfo{pages}{1264--1272}.
\newblock \DOIprefix\doi{10.1016/j.ophtha.2018.01.034}.
\bibitem[{Lehmann and Romano(2022)}]{lehmann2022testing}
\bibinfo{author}{Lehmann, E.L.}, \bibinfo{author}{Romano, J.P.},
  \bibinfo{year}{2022}.
\newblock \bibinfo{title}{Testing Statistical Hypotheses}.
\newblock Springer Texts in Statistics. \bibinfo{edition}{4th} ed.,
  \bibinfo{publisher}{Springer}, \bibinfo{address}{Cham}.
\newblock \DOIprefix\doi{10.1007/978-3-030-70578-7}.
\bibitem[{Leibig et~al.(2017)Leibig, Allken, Ayhan, Berens and
  Wahl}]{leibig2017leveraging}
\bibinfo{author}{Leibig, C.}, \bibinfo{author}{Allken, V.},
  \bibinfo{author}{Ayhan, M.S.}, \bibinfo{author}{Berens, P.},
  \bibinfo{author}{Wahl, S.}, \bibinfo{year}{2017}.
\newblock \bibinfo{title}{Leveraging uncertainty information from deep neural
  networks for disease detection}.
\newblock \bibinfo{journal}{Sci. Rep.} \bibinfo{volume}{7},
  \bibinfo{pages}{17816}.
\newblock \DOIprefix\doi{10.1038/s41598-017-17876-z}.
\bibitem[{Loshchilov and Hutter(2019)}]{loshchilov2019decoupled}
\bibinfo{author}{Loshchilov, I.}, \bibinfo{author}{Hutter, F.},
  \bibinfo{year}{2019}.
\newblock \bibinfo{title}{Decoupled weight decay regularization}, in:
  \bibinfo{booktitle}{Seventh International Conference on Learning
  Representations ({ICLR} 2019)}.
\newblock \URLprefix \url{https://openreview.net/forum?id=Bkg6RiCqY7}.
\bibitem[{Lu et~al.(2022)Lu, Lemay, Chang, H{\"o}bel and
  Kalpathy-Cramer}]{lu2022fair}
\bibinfo{author}{Lu, C.}, \bibinfo{author}{Lemay, A.}, \bibinfo{author}{Chang,
  K.}, \bibinfo{author}{H{\"o}bel, K.}, \bibinfo{author}{Kalpathy-Cramer, J.},
  \bibinfo{year}{2022}.
\newblock \bibinfo{title}{Fair conformal predictors for applications in medical
  imaging}, in: \bibinfo{booktitle}{Proceedings of the AAAI Conference on
  Artificial Intelligence}, pp. \bibinfo{pages}{12008--12016}.
\newblock \DOIprefix\doi{10.1609/aaai.v36i11.21459}.
\bibitem[{Maier-Hein et~al.(2018)Maier-Hein, Eisenmann, Reinke, Onogur,
  Stankovic, Scholz, Arbel, Bogunovic, Bradley, Carass, Feldmann, Frangi, Full,
  van Ginneken, Hanbury, Honauer, Kozubek, Landman, M{\"a}rz, Maier,
  Maier-Hein, Menze, M{\"u}ller, Neher, Niessen, Rajpoot, Sharp,
  Sirinukunwattana, Speidel, Stock, Stoyanov, Taha, van~der Sommen, Wang,
  Weber, Zheng, Jannin and Kopp-Schneider}]{maierhein2018rankings}
\bibinfo{author}{Maier-Hein, L.}, \bibinfo{author}{Eisenmann, M.},
  \bibinfo{author}{Reinke, A.}, \bibinfo{author}{Onogur, S.},
  \bibinfo{author}{Stankovic, M.}, \bibinfo{author}{Scholz, P.},
  \bibinfo{author}{Arbel, T.}, \bibinfo{author}{Bogunovic, H.},
  \bibinfo{author}{Bradley, A.P.}, \bibinfo{author}{Carass, A.},
  \bibinfo{author}{Feldmann, C.}, \bibinfo{author}{Frangi, A.F.},
  \bibinfo{author}{Full, P.M.}, \bibinfo{author}{van Ginneken, B.},
  \bibinfo{author}{Hanbury, A.}, \bibinfo{author}{Honauer, K.},
  \bibinfo{author}{Kozubek, M.}, \bibinfo{author}{Landman, B.A.},
  \bibinfo{author}{M{\"a}rz, K.}, \bibinfo{author}{Maier, O.},
  \bibinfo{author}{Maier-Hein, K.}, \bibinfo{author}{Menze, B.H.},
  \bibinfo{author}{M{\"u}ller, H.}, \bibinfo{author}{Neher, P.F.},
  \bibinfo{author}{Niessen, W.}, \bibinfo{author}{Rajpoot, N.},
  \bibinfo{author}{Sharp, G.C.}, \bibinfo{author}{Sirinukunwattana, K.},
  \bibinfo{author}{Speidel, S.}, \bibinfo{author}{Stock, C.},
  \bibinfo{author}{Stoyanov, D.}, \bibinfo{author}{Taha, A.A.},
  \bibinfo{author}{van~der Sommen, F.}, \bibinfo{author}{Wang, C.W.},
  \bibinfo{author}{Weber, M.A.}, \bibinfo{author}{Zheng, G.},
  \bibinfo{author}{Jannin, P.}, \bibinfo{author}{Kopp-Schneider, A.},
  \bibinfo{year}{2018}.
\newblock \bibinfo{title}{Why rankings of biomedical image analysis
  competitions should be interpreted with care}.
\newblock \bibinfo{journal}{Nat. Commun.} \bibinfo{volume}{9},
  \bibinfo{pages}{5217}.
\newblock \DOIprefix\doi{10.1038/s41467-018-07619-7}.
\bibitem[{Maier-Hein et~al.(2024)Maier-Hein, Reinke, Godau, Tizabi, Buettner,
  Christodoulou, Glocker, Isensee, Kleesiek, Kozubek, Reyes, Riegler,
  Wiesenfarth, Kavur, Sudre, Baumgartner, Eisenmann, Heckmann-N{\"o}tzel,
  R{\"a}dsch, Acion, Antonelli, Arbel, Bakas, Benis, Blaschko, Cardoso,
  Cheplygina, Cimini, Collins, Farahani, Ferrer, Galdran, van Ginneken, Haase,
  Hashimoto, Hoffman, Huisman, Jannin, Kahn, Kainmueller, Kainz, Karargyris,
  Karthikesalingam, Kofler, Kopp-Schneider, Kreshuk, Kurc, Landman, Litjens,
  Madani, Maier-Hein, Martel, Mattson, Meijering, Menze, Moons, M{\"u}ller,
  Nichyporuk, Nickel, Petersen, Rajpoot, Rieke, Saez-Rodriguez, S{\'a}nchez,
  Shetty, van Smeden, Summers, Taha, Tiulpin, Tsaftaris, Van~Calster, Varoquaux
  and J{\"a}ger}]{maierhein2024metrics}
\bibinfo{author}{Maier-Hein, L.}, \bibinfo{author}{Reinke, A.},
  \bibinfo{author}{Godau, P.}, \bibinfo{author}{Tizabi, M.D.},
  \bibinfo{author}{Buettner, F.}, \bibinfo{author}{Christodoulou, E.},
  \bibinfo{author}{Glocker, B.}, \bibinfo{author}{Isensee, F.},
  \bibinfo{author}{Kleesiek, J.}, \bibinfo{author}{Kozubek, M.},
  \bibinfo{author}{Reyes, M.}, \bibinfo{author}{Riegler, M.A.},
  \bibinfo{author}{Wiesenfarth, M.}, \bibinfo{author}{Kavur, A.E.},
  \bibinfo{author}{Sudre, C.H.}, \bibinfo{author}{Baumgartner, M.},
  \bibinfo{author}{Eisenmann, M.}, \bibinfo{author}{Heckmann-N{\"o}tzel, D.},
  \bibinfo{author}{R{\"a}dsch, T.}, \bibinfo{author}{Acion, L.},
  \bibinfo{author}{Antonelli, M.}, \bibinfo{author}{Arbel, T.},
  \bibinfo{author}{Bakas, S.}, \bibinfo{author}{Benis, A.},
  \bibinfo{author}{Blaschko, M.B.}, \bibinfo{author}{Cardoso, M.J.},
  \bibinfo{author}{Cheplygina, V.}, \bibinfo{author}{Cimini, B.A.},
  \bibinfo{author}{Collins, G.S.}, \bibinfo{author}{Farahani, K.},
  \bibinfo{author}{Ferrer, L.}, \bibinfo{author}{Galdran, A.},
  \bibinfo{author}{van Ginneken, B.}, \bibinfo{author}{Haase, R.},
  \bibinfo{author}{Hashimoto, D.A.}, \bibinfo{author}{Hoffman, M.M.},
  \bibinfo{author}{Huisman, M.}, \bibinfo{author}{Jannin, P.},
  \bibinfo{author}{Kahn, C.E.}, \bibinfo{author}{Kainmueller, D.},
  \bibinfo{author}{Kainz, B.}, \bibinfo{author}{Karargyris, A.},
  \bibinfo{author}{Karthikesalingam, A.}, \bibinfo{author}{Kofler, F.},
  \bibinfo{author}{Kopp-Schneider, A.}, \bibinfo{author}{Kreshuk, A.},
  \bibinfo{author}{Kurc, T.}, \bibinfo{author}{Landman, B.A.},
  \bibinfo{author}{Litjens, G.}, \bibinfo{author}{Madani, A.},
  \bibinfo{author}{Maier-Hein, K.}, \bibinfo{author}{Martel, A.L.},
  \bibinfo{author}{Mattson, P.}, \bibinfo{author}{Meijering, E.},
  \bibinfo{author}{Menze, B.}, \bibinfo{author}{Moons, K.G.M.},
  \bibinfo{author}{M{\"u}ller, H.}, \bibinfo{author}{Nichyporuk, B.},
  \bibinfo{author}{Nickel, F.}, \bibinfo{author}{Petersen, J.},
  \bibinfo{author}{Rajpoot, N.}, \bibinfo{author}{Rieke, N.},
  \bibinfo{author}{Saez-Rodriguez, J.}, \bibinfo{author}{S{\'a}nchez, C.I.},
  \bibinfo{author}{Shetty, S.}, \bibinfo{author}{van Smeden, M.},
  \bibinfo{author}{Summers, R.M.}, \bibinfo{author}{Taha, A.A.},
  \bibinfo{author}{Tiulpin, A.}, \bibinfo{author}{Tsaftaris, S.A.},
  \bibinfo{author}{Van~Calster, B.}, \bibinfo{author}{Varoquaux, G.},
  \bibinfo{author}{J{\"a}ger, P.F.}, \bibinfo{year}{2024}.
\newblock \bibinfo{title}{Metrics reloaded: recommendations for image analysis
  validation}.
\newblock \bibinfo{journal}{Nat. Methods} \bibinfo{volume}{21},
  \bibinfo{pages}{195--212}.
\newblock \DOIprefix\doi{10.1038/s41592-023-02151-z}.
\bibitem[{Maier-Hein et~al.(2017)Maier-Hein, Vedula, Speidel, Navab, Kikinis,
  Park, Eisenmann, Feussner, Forestier, Giannarou, Hashizume, Katic, Kenngott,
  Kranzfelder, Malpani, M{\"a}rz, Neumuth, Padoy, Pugh, Schoch, Stoyanov,
  Taylor, Wagner, Hager and Jannin}]{maierhein2017surgical}
\bibinfo{author}{Maier-Hein, L.}, \bibinfo{author}{Vedula, S.S.},
  \bibinfo{author}{Speidel, S.}, \bibinfo{author}{Navab, N.},
  \bibinfo{author}{Kikinis, R.}, \bibinfo{author}{Park, A.},
  \bibinfo{author}{Eisenmann, M.}, \bibinfo{author}{Feussner, H.},
  \bibinfo{author}{Forestier, G.}, \bibinfo{author}{Giannarou, S.},
  \bibinfo{author}{Hashizume, M.}, \bibinfo{author}{Katic, D.},
  \bibinfo{author}{Kenngott, H.}, \bibinfo{author}{Kranzfelder, M.},
  \bibinfo{author}{Malpani, A.}, \bibinfo{author}{M{\"a}rz, K.},
  \bibinfo{author}{Neumuth, T.}, \bibinfo{author}{Padoy, N.},
  \bibinfo{author}{Pugh, C.}, \bibinfo{author}{Schoch, N.},
  \bibinfo{author}{Stoyanov, D.}, \bibinfo{author}{Taylor, R.},
  \bibinfo{author}{Wagner, M.}, \bibinfo{author}{Hager, G.D.},
  \bibinfo{author}{Jannin, P.}, \bibinfo{year}{2017}.
\newblock \bibinfo{title}{Surgical data science for next-generation
  interventions}.
\newblock \bibinfo{journal}{Nat. Biomed. Eng.} \bibinfo{volume}{1},
  \bibinfo{pages}{691--696}.
\newblock \DOIprefix\doi{10.1038/s41551-017-0132-7}.
\bibitem[{Mascagni et~al.(2021)Mascagni, Alapatt, Garcia, Okamoto, Vardazaryan,
  Costamagna, Dallemagne and Padoy}]{mascagni2021protocol}
\bibinfo{author}{Mascagni, P.}, \bibinfo{author}{Alapatt, D.},
  \bibinfo{author}{Garcia, A.}, \bibinfo{author}{Okamoto, N.},
  \bibinfo{author}{Vardazaryan, A.}, \bibinfo{author}{Costamagna, G.},
  \bibinfo{author}{Dallemagne, B.}, \bibinfo{author}{Padoy, N.},
  \bibinfo{year}{2021}.
\newblock \bibinfo{title}{Surgical data science for safe cholecystectomy: a
  protocol for segmentation of hepatocystic anatomy and assessment of the
  critical view of safety}.
\newblock \bibinfo{howpublished}{arXiv preprint arXiv:2106.10916}.
\newblock \DOIprefix\doi{10.48550/arXiv.2106.10916}.
\bibitem[{Mascagni et~al.(2025)Mascagni, Alapatt, Murali, Vardazaryan, Garcia,
  Okamoto, Costamagna, Mutter, Marescaux, Dallemagne and
  Padoy}]{murali2023endoscapes}
\bibinfo{author}{Mascagni, P.}, \bibinfo{author}{Alapatt, D.},
  \bibinfo{author}{Murali, A.}, \bibinfo{author}{Vardazaryan, A.},
  \bibinfo{author}{Garcia, A.}, \bibinfo{author}{Okamoto, N.},
  \bibinfo{author}{Costamagna, G.}, \bibinfo{author}{Mutter, D.},
  \bibinfo{author}{Marescaux, J.}, \bibinfo{author}{Dallemagne, B.},
  \bibinfo{author}{Padoy, N.}, \bibinfo{year}{2025}.
\newblock \bibinfo{title}{{Endoscapes}, a critical view of safety and surgical
  scene segmentation dataset for laparoscopic cholecystectomy}.
\newblock \bibinfo{journal}{Sci. Data} \bibinfo{volume}{12},
  \bibinfo{pages}{331}.
\newblock \DOIprefix\doi{10.1038/s41597-025-04642-4}.
\bibitem[{Mascagni et~al.(2022)Mascagni, Vardazaryan, Alapatt, Urade, Emre,
  Fiorillo, Pessaux, Mutter, Marescaux, Costamagna, Dallemagne and
  Padoy}]{mascagni2022artificial}
\bibinfo{author}{Mascagni, P.}, \bibinfo{author}{Vardazaryan, A.},
  \bibinfo{author}{Alapatt, D.}, \bibinfo{author}{Urade, T.},
  \bibinfo{author}{Emre, T.}, \bibinfo{author}{Fiorillo, C.},
  \bibinfo{author}{Pessaux, P.}, \bibinfo{author}{Mutter, D.},
  \bibinfo{author}{Marescaux, J.}, \bibinfo{author}{Costamagna, G.},
  \bibinfo{author}{Dallemagne, B.}, \bibinfo{author}{Padoy, N.},
  \bibinfo{year}{2022}.
\newblock \bibinfo{title}{Artificial intelligence for surgical safety:
  automatic assessment of the critical view of safety in laparoscopic
  cholecystectomy using deep learning}.
\newblock \bibinfo{journal}{Ann. Surg.} \bibinfo{volume}{275},
  \bibinfo{pages}{955--961}.
\newblock \DOIprefix\doi{10.1097/SLA.0000000000004351}.
\bibitem[{McNemar(1947)}]{mcnemar1947note}
\bibinfo{author}{McNemar, Q.}, \bibinfo{year}{1947}.
\newblock \bibinfo{title}{Note on the sampling error of the difference between
  correlated proportions or percentages}.
\newblock \bibinfo{journal}{Psychometrika} \bibinfo{volume}{12},
  \bibinfo{pages}{153--157}.
\newblock \DOIprefix\doi{10.1007/BF02295996}.
\bibitem[{Murali et~al.(2024)Murali, Alapatt, Mascagni, Vardazaryan, Garcia,
  Okamoto, Mutter and Padoy}]{murali2023latent}
\bibinfo{author}{Murali, A.}, \bibinfo{author}{Alapatt, D.},
  \bibinfo{author}{Mascagni, P.}, \bibinfo{author}{Vardazaryan, A.},
  \bibinfo{author}{Garcia, A.}, \bibinfo{author}{Okamoto, N.},
  \bibinfo{author}{Mutter, D.}, \bibinfo{author}{Padoy, N.},
  \bibinfo{year}{2024}.
\newblock \bibinfo{title}{Latent graph representations for critical view of
  safety assessment}.
\newblock \bibinfo{journal}{IEEE Trans. Med. Imaging} \bibinfo{volume}{43},
  \bibinfo{pages}{1247--1258}.
\newblock \DOIprefix\doi{10.1109/TMI.2023.3333034}.
\bibitem[{Neyman and Pearson(1933)}]{neyman1933problem}
\bibinfo{author}{Neyman, J.}, \bibinfo{author}{Pearson, E.S.},
  \bibinfo{year}{1933}.
\newblock \bibinfo{title}{On the problem of the most efficient tests of
  statistical hypotheses}.
\newblock \bibinfo{journal}{Philos. Trans. R. Soc. Lond. Ser. A}
  \bibinfo{volume}{231}, \bibinfo{pages}{289--337}.
\newblock \DOIprefix\doi{10.1098/rsta.1933.0009}.
\bibitem[{Nijssen et~al.(2015)Nijssen, Schreinemakers, Meyer, van~der
  Schelling, Crolla and Rijken}]{nijssen2015complications}
\bibinfo{author}{Nijssen, M.A.J.}, \bibinfo{author}{Schreinemakers, J.M.J.},
  \bibinfo{author}{Meyer, Z.}, \bibinfo{author}{van~der Schelling, G.P.},
  \bibinfo{author}{Crolla, R.M.P.H.}, \bibinfo{author}{Rijken, A.M.},
  \bibinfo{year}{2015}.
\newblock \bibinfo{title}{Complications after laparoscopic cholecystectomy: a
  video evaluation study of whether the critical view of safety was reached}.
\newblock \bibinfo{journal}{World J. Surg.} \bibinfo{volume}{39},
  \bibinfo{pages}{1798--1803}.
\newblock \DOIprefix\doi{10.1007/s00268-015-2993-9}.
\bibitem[{Oakden-Rayner et~al.(2020)Oakden-Rayner, Dunnmon, Carneiro and
  R{\'e}}]{oakdenrayner2020hidden}
\bibinfo{author}{Oakden-Rayner, L.}, \bibinfo{author}{Dunnmon, J.},
  \bibinfo{author}{Carneiro, G.}, \bibinfo{author}{R{\'e}, C.},
  \bibinfo{year}{2020}.
\newblock \bibinfo{title}{Hidden stratification causes clinically meaningful
  failures in machine learning for medical imaging}, in:
  \bibinfo{booktitle}{Proceedings of the ACM Conference on Health, Inference,
  and Learning ({CHIL} '20)}, \bibinfo{publisher}{Association for Computing
  Machinery}, \bibinfo{address}{New York, NY}. pp. \bibinfo{pages}{151--159}.
\newblock \DOIprefix\doi{10.1145/3368555.3384468}.
\bibitem[{Page(1954)}]{page1954continuous}
\bibinfo{author}{Page, E.S.}, \bibinfo{year}{1954}.
\newblock \bibinfo{title}{Continuous inspection schemes}.
\newblock \bibinfo{journal}{Biometrika} \bibinfo{volume}{41},
  \bibinfo{pages}{100--115}.
\newblock \DOIprefix\doi{10.1093/biomet/41.1-2.100}.
\bibitem[{Park et~al.(2020)Park, Bastani, Matni and Lee}]{park2020pac}
\bibinfo{author}{Park, S.}, \bibinfo{author}{Bastani, O.},
  \bibinfo{author}{Matni, N.}, \bibinfo{author}{Lee, I.}, \bibinfo{year}{2020}.
\newblock \bibinfo{title}{{PAC} confidence sets for deep neural networks via
  calibrated prediction}, in: \bibinfo{booktitle}{Eighth International
  Conference on Learning Representations ({ICLR} 2020)}.
\newblock \URLprefix \url{https://openreview.net/forum?id=BJxVI04YvB}.
\bibitem[{Rasch(1960)}]{rasch1960probabilistic}
\bibinfo{author}{Rasch, G.}, \bibinfo{year}{1960}.
\newblock \bibinfo{title}{Probabilistic models for some intelligence and
  attainment tests}. volume~\bibinfo{volume}{1} of
  \textit{\bibinfo{series}{Studies in Mathematical Psychology}}.
\newblock \bibinfo{publisher}{Danish Institute for Educational Research},
  \bibinfo{address}{Copenhagen}.
\bibitem[{Raykar et~al.(2010)Raykar, Yu, Zhao, Valadez, Florin, Bogoni and
  Moy}]{raykar2010learning}
\bibinfo{author}{Raykar, V.C.}, \bibinfo{author}{Yu, S.},
  \bibinfo{author}{Zhao, L.H.}, \bibinfo{author}{Valadez, G.H.},
  \bibinfo{author}{Florin, C.}, \bibinfo{author}{Bogoni, L.},
  \bibinfo{author}{Moy, L.}, \bibinfo{year}{2010}.
\newblock \bibinfo{title}{Learning from crowds}.
\newblock \bibinfo{journal}{J. Mach. Learn. Res.} \bibinfo{volume}{11},
  \bibinfo{pages}{1297--1322}.
\newblock \URLprefix \url{https://jmlr.org/papers/v11/raykar10a.html}.
\bibitem[{Reitsma et~al.(2009)Reitsma, Rutjes, Khan, Coomarasamy and
  Bossuyt}]{reitsma2009review}
\bibinfo{author}{Reitsma, J.B.}, \bibinfo{author}{Rutjes, A.W.S.},
  \bibinfo{author}{Khan, K.S.}, \bibinfo{author}{Coomarasamy, A.},
  \bibinfo{author}{Bossuyt, P.M.}, \bibinfo{year}{2009}.
\newblock \bibinfo{title}{A review of solutions for diagnostic accuracy studies
  with an imperfect or missing reference standard}.
\newblock \bibinfo{journal}{J. Clin. Epidemiol.} \bibinfo{volume}{62},
  \bibinfo{pages}{797--806}.
\newblock \DOIprefix\doi{10.1016/j.jclinepi.2009.02.005}.
\bibitem[{R{\'\i}os et~al.(2023)R{\'\i}os, Molina-Rodriguez, Londo{\~n}o,
  Guill{\'e}n, Sierra, Zapata and Giraldo}]{rios2023cholec80cvs}
\bibinfo{author}{R{\'\i}os, M.S.}, \bibinfo{author}{Molina-Rodriguez, M.A.},
  \bibinfo{author}{Londo{\~n}o, D.}, \bibinfo{author}{Guill{\'e}n, C.A.},
  \bibinfo{author}{Sierra, S.}, \bibinfo{author}{Zapata, F.},
  \bibinfo{author}{Giraldo, L.F.}, \bibinfo{year}{2023}.
\newblock \bibinfo{title}{{Cholec80-CVS}: an open dataset with an evaluation of
  {Strasberg}'s critical view of safety for {AI}}.
\newblock \bibinfo{journal}{Sci. Data} \bibinfo{volume}{10},
  \bibinfo{pages}{194}.
\newblock \DOIprefix\doi{10.1038/s41597-023-02073-7}.
\bibitem[{Romano et~al.(2020)Romano, Barber, Sabatti and
  Cand{\`e}s}]{romano2020malice}
\bibinfo{author}{Romano, Y.}, \bibinfo{author}{Barber, R.F.},
  \bibinfo{author}{Sabatti, C.}, \bibinfo{author}{Cand{\`e}s, E.},
  \bibinfo{year}{2020}.
\newblock \bibinfo{title}{With malice toward none: assessing uncertainty via
  equalized coverage}.
\newblock \bibinfo{journal}{Harv. Data Sci. Rev.} \bibinfo{volume}{2}.
\newblock \DOIprefix\doi{10.1162/99608f92.03f00592}.
\bibitem[{Rystedt et~al.(2016)Rystedt, Lindell and
  Montgomery}]{rystedt2016bile}
\bibinfo{author}{Rystedt, J.}, \bibinfo{author}{Lindell, G.},
  \bibinfo{author}{Montgomery, A.}, \bibinfo{year}{2016}.
\newblock \bibinfo{title}{Bile duct injuries associated with 55,134
  cholecystectomies: treatment and outcome from a national perspective}.
\newblock \bibinfo{journal}{World J. Surg.} \bibinfo{volume}{40},
  \bibinfo{pages}{73--80}.
\newblock \DOIprefix\doi{10.1007/s00268-015-3281-4}.
\bibitem[{Sim{\'e}oni et~al.(2025)Sim{\'e}oni, Vo, Seitzer, Baldassarre, Oquab,
  Jose, Khalidov, Szafraniec, Yi, Ramamonjisoa, Massa, Haziza, Wehrstedt, Wang,
  Darcet, Moutakanni, Sentana, Roberts, Vedaldi, Tolan, Brandt, Couprie,
  Mairal, J{\'e}gou, Labatut and Bojanowski}]{simeoni2025dinov3}
\bibinfo{author}{Sim{\'e}oni, O.}, \bibinfo{author}{Vo, H.V.},
  \bibinfo{author}{Seitzer, M.}, \bibinfo{author}{Baldassarre, F.},
  \bibinfo{author}{Oquab, M.}, \bibinfo{author}{Jose, C.},
  \bibinfo{author}{Khalidov, V.}, \bibinfo{author}{Szafraniec, M.},
  \bibinfo{author}{Yi, S.}, \bibinfo{author}{Ramamonjisoa, M.},
  \bibinfo{author}{Massa, F.}, \bibinfo{author}{Haziza, D.},
  \bibinfo{author}{Wehrstedt, L.}, \bibinfo{author}{Wang, J.},
  \bibinfo{author}{Darcet, T.}, \bibinfo{author}{Moutakanni, T.},
  \bibinfo{author}{Sentana, L.}, \bibinfo{author}{Roberts, C.},
  \bibinfo{author}{Vedaldi, A.}, \bibinfo{author}{Tolan, J.},
  \bibinfo{author}{Brandt, J.}, \bibinfo{author}{Couprie, C.},
  \bibinfo{author}{Mairal, J.}, \bibinfo{author}{J{\'e}gou, H.},
  \bibinfo{author}{Labatut, P.}, \bibinfo{author}{Bojanowski, P.},
  \bibinfo{year}{2025}.
\newblock \bibinfo{title}{{DINOv3}}.
\newblock \bibinfo{howpublished}{arXiv preprint arXiv:2508.10104}.
\newblock \DOIprefix\doi{10.48550/arXiv.2508.10104}.
\bibitem[{Sohoni et~al.(2020)Sohoni, Dunnmon, Angus, Gu and
  R{\'e}}]{sohoni2020nosubclass}
\bibinfo{author}{Sohoni, N.S.}, \bibinfo{author}{Dunnmon, J.A.},
  \bibinfo{author}{Angus, G.}, \bibinfo{author}{Gu, A.},
  \bibinfo{author}{R{\'e}, C.}, \bibinfo{year}{2020}.
\newblock \bibinfo{title}{No subclass left behind: fine-grained robustness in
  coarse-grained classification problems}, in: \bibinfo{booktitle}{Advances in
  Neural Information Processing Systems}, \bibinfo{publisher}{Curran
  Associates}. pp. \bibinfo{pages}{19339--19352}.
\newblock \URLprefix
  \url{https://proceedings.neurips.cc/paper_files/paper/2020/hash/e0688d13958a19e087e123148555e4b4-Abstract.html}.
\bibitem[{Strasberg and Brunt(2010)}]{strasberg2010rationale}
\bibinfo{author}{Strasberg, S.M.}, \bibinfo{author}{Brunt, L.M.},
  \bibinfo{year}{2010}.
\newblock \bibinfo{title}{Rationale and use of the critical view of safety in
  laparoscopic cholecystectomy}.
\newblock \bibinfo{journal}{J. Am. Coll. Surg.} \bibinfo{volume}{211},
  \bibinfo{pages}{132--138}.
\newblock \DOIprefix\doi{10.1016/j.jamcollsurg.2010.02.053}.
\bibitem[{Strasberg et~al.(1995)Strasberg, Hertl and
  Soper}]{strasberg1995analysis}
\bibinfo{author}{Strasberg, S.M.}, \bibinfo{author}{Hertl, M.},
  \bibinfo{author}{Soper, N.J.}, \bibinfo{year}{1995}.
\newblock \bibinfo{title}{An analysis of the problem of biliary injury during
  laparoscopic cholecystectomy}.
\newblock \bibinfo{journal}{J. Am. Coll. Surg.} \bibinfo{volume}{180},
  \bibinfo{pages}{101--125}.
\newblock \bibinfo{note}{PMID: 8000648}.
\bibitem[{Stutz et~al.(2025)Stutz, Cemgil, Roy, Matejovicova, Barsbey,
  Strachan, Schaekermann, Freyberg, Rikhye, Freeman, Matos, Telang, Webster,
  Liu, Corrado, Matias, Kohli, Liu, Doucet and
  Karthikesalingam}]{stutz2023evaluating}
\bibinfo{author}{Stutz, D.}, \bibinfo{author}{Cemgil, A.T.},
  \bibinfo{author}{Roy, A.G.}, \bibinfo{author}{Matejovicova, T.},
  \bibinfo{author}{Barsbey, M.}, \bibinfo{author}{Strachan, P.},
  \bibinfo{author}{Schaekermann, M.}, \bibinfo{author}{Freyberg, J.},
  \bibinfo{author}{Rikhye, R.}, \bibinfo{author}{Freeman, B.},
  \bibinfo{author}{Matos, J.P.}, \bibinfo{author}{Telang, U.},
  \bibinfo{author}{Webster, D.R.}, \bibinfo{author}{Liu, Y.},
  \bibinfo{author}{Corrado, G.S.}, \bibinfo{author}{Matias, Y.},
  \bibinfo{author}{Kohli, P.}, \bibinfo{author}{Liu, Y.},
  \bibinfo{author}{Doucet, A.}, \bibinfo{author}{Karthikesalingam, A.},
  \bibinfo{year}{2025}.
\newblock \bibinfo{title}{Evaluating medical {AI} systems in dermatology under
  uncertain ground truth}.
\newblock \bibinfo{journal}{Med. Image Anal.} \bibinfo{volume}{103},
  \bibinfo{pages}{103556}.
\newblock \DOIprefix\doi{10.1016/j.media.2025.103556}.
\bibitem[{Stutz et~al.(2023)Stutz, Roy, Matejovicova, Strachan, Cemgil and
  Doucet}]{stutz2023conformal}
\bibinfo{author}{Stutz, D.}, \bibinfo{author}{Roy, A.G.},
  \bibinfo{author}{Matejovicova, T.}, \bibinfo{author}{Strachan, P.},
  \bibinfo{author}{Cemgil, A.T.}, \bibinfo{author}{Doucet, A.},
  \bibinfo{year}{2023}.
\newblock \bibinfo{title}{Conformal prediction under ambiguous ground truth}.
\newblock \bibinfo{journal}{Trans. Mach. Learn. Res.} \URLprefix
  \url{https://openreview.net/forum?id=CAd6V2qXxc}.
\bibitem[{Tong et~al.(2018)Tong, Feng and Li}]{tong2018neyman}
\bibinfo{author}{Tong, X.}, \bibinfo{author}{Feng, Y.}, \bibinfo{author}{Li,
  J.J.}, \bibinfo{year}{2018}.
\newblock \bibinfo{title}{{Neyman-Pearson} classification algorithms and {NP}
  receiver operating characteristics}.
\newblock \bibinfo{journal}{Sci. Adv.} \bibinfo{volume}{4},
  \bibinfo{pages}{eaao1659}.
\newblock \DOIprefix\doi{10.1126/sciadv.aao1659}.
\bibitem[{Twinanda et~al.(2017)Twinanda, Shehata, Mutter, Marescaux,
  de~Mathelin and Padoy}]{twinanda2017endonet}
\bibinfo{author}{Twinanda, A.P.}, \bibinfo{author}{Shehata, S.},
  \bibinfo{author}{Mutter, D.}, \bibinfo{author}{Marescaux, J.},
  \bibinfo{author}{de~Mathelin, M.}, \bibinfo{author}{Padoy, N.},
  \bibinfo{year}{2017}.
\newblock \bibinfo{title}{{EndoNet}: a deep architecture for recognition tasks
  on laparoscopic videos}.
\newblock \bibinfo{journal}{IEEE Trans. Med. Imaging} \bibinfo{volume}{36},
  \bibinfo{pages}{86--97}.
\newblock \DOIprefix\doi{10.1109/TMI.2016.2593957}.
\bibitem[{Varoquaux and Cheplygina(2022)}]{varoquaux2022machine}
\bibinfo{author}{Varoquaux, G.}, \bibinfo{author}{Cheplygina, V.},
  \bibinfo{year}{2022}.
\newblock \bibinfo{title}{Machine learning for medical imaging: methodological
  failures and recommendations for the future}.
\newblock \bibinfo{journal}{npj Digit. Med.} \bibinfo{volume}{5},
  \bibinfo{pages}{48}.
\newblock \DOIprefix\doi{10.1038/s41746-022-00592-y}.
\bibitem[{Vovk(2012)}]{vovk2012conditional}
\bibinfo{author}{Vovk, V.}, \bibinfo{year}{2012}.
\newblock \bibinfo{title}{Conditional validity of inductive conformal
  predictors}, in: \bibinfo{booktitle}{Proceedings of the Asian Conference on
  Machine Learning ({ACML})}, \bibinfo{publisher}{PMLR}. pp.
  \bibinfo{pages}{475--490}.
\newblock \URLprefix \url{https://proceedings.mlr.press/v25/vovk12.html}.
\bibitem[{Vovk(2015)}]{vovk2015cross}
\bibinfo{author}{Vovk, V.}, \bibinfo{year}{2015}.
\newblock \bibinfo{title}{Cross-conformal predictors}.
\newblock \bibinfo{journal}{Ann. Math. Artif. Intell.} \bibinfo{volume}{74},
  \bibinfo{pages}{9--28}.
\newblock \DOIprefix\doi{10.1007/s10472-013-9368-4}.
\bibitem[{Vovk et~al.(2005)Vovk, Gammerman and Shafer}]{vovk2005algorithmic}
\bibinfo{author}{Vovk, V.}, \bibinfo{author}{Gammerman, A.},
  \bibinfo{author}{Shafer, G.}, \bibinfo{year}{2005}.
\newblock \bibinfo{title}{Algorithmic Learning in a Random World}.
\newblock \bibinfo{edition}{1st} ed., \bibinfo{publisher}{Springer},
  \bibinfo{address}{New York}.
\newblock \DOIprefix\doi{10.1007/b106715}.
\bibitem[{Way et~al.(2003)Way, Stewart, Gantert, Liu, Lee, Whang and
  Hunter}]{way2003causes}
\bibinfo{author}{Way, L.W.}, \bibinfo{author}{Stewart, L.},
  \bibinfo{author}{Gantert, W.}, \bibinfo{author}{Liu, K.},
  \bibinfo{author}{Lee, C.M.}, \bibinfo{author}{Whang, K.},
  \bibinfo{author}{Hunter, J.G.}, \bibinfo{year}{2003}.
\newblock \bibinfo{title}{Causes and prevention of laparoscopic bile duct
  injuries: analysis of 252 cases from a human factors and cognitive psychology
  perspective}.
\newblock \bibinfo{journal}{Ann. Surg.} \bibinfo{volume}{237},
  \bibinfo{pages}{460--469}.
\newblock \DOIprefix\doi{10.1097/01.SLA.0000060680.92690.E9}.
\bibitem[{Zhang et~al.(2024)Zhang, Roth and Zhang}]{zhang2024fair}
\bibinfo{author}{Zhang, L.}, \bibinfo{author}{Roth, A.},
  \bibinfo{author}{Zhang, L.}, \bibinfo{year}{2024}.
\newblock \bibinfo{title}{Fair risk control: a generalized framework for
  calibrating multi-group fairness risks}, in: \bibinfo{booktitle}{Proceedings
  of the 41st International Conference on Machine Learning ({ICML})},
  \bibinfo{publisher}{PMLR}. pp. \bibinfo{pages}{59783--59805}.
\newblock \URLprefix \url{https://proceedings.mlr.press/v235/zhang24be.html}.

\end{thebibliography}

\end{document}